\documentclass{article}

\usepackage{microtype}
\usepackage{graphicx}
\usepackage{subcaption}
\usepackage{booktabs} 

\usepackage{hyperref}

\usepackage[accepted]{icml2026}

\usepackage{amsmath}
\usepackage{amssymb}
\usepackage{mathtools}
\usepackage{amsthm}
\usepackage{algorithm}
\usepackage{algorithmic}
\usepackage{multirow}

\usepackage[capitalize,noabbrev]{cleveref}

\theoremstyle{plain}
\newtheorem{theorem}{Theorem}[section]
\newtheorem{proposition}[theorem]{Proposition}
\newtheorem{lemma}[theorem]{Lemma}

\theoremstyle{definition}
\newtheorem{definition}[theorem]{Definition}

\theoremstyle{remark}

\usepackage{tikz}
\usepackage{pgfplots}
\pgfplotsset{compat=1.17}
\usetikzlibrary{patterns, arrows.meta, calc, backgrounds, decorations.pathmorphing, shadows,patterns.meta, intersections}
\usetikzlibrary{positioning, fit, shapes, arrows.meta, shadows, patterns.meta, shapes.geometric}
\usepgfplotslibrary{fillbetween}
\definecolor{icmlblue}{RGB}{59, 91, 167}
\definecolor{icmlred}{RGB}{209, 73, 91} 
\definecolor{icmlgray}{RGB}{100, 100, 100}
\definecolor{icmlorange}{RGB}{230, 159, 0} 
\definecolor{icmlblack}{RGB}{255, 255, 255} 
\usepackage{listings}
\usepackage{xcolor}

\definecolor{codegreen}{rgb}{0,0.5,0}
\definecolor{codegray}{rgb}{0.5,0.5,0.5}
\definecolor{codepurple}{rgb}{0.58,0,0.82}
\definecolor{backcolour}{rgb}{0.97,0.97,0.97}
\definecolor{deepblue}{rgb}{0,0,0.5}
\definecolor{deepred}{rgb}{0.6,0,0}

\lstdefinestyle{icmlstyle}{
    backgroundcolor=\color{backcolour},   
    commentstyle=\color{codegreen},
    keywordstyle=\color{deepblue}\bfseries,
    numberstyle=\tiny\color{codegray},
    stringstyle=\color{deepred},
    basicstyle=\ttfamily\footnotesize,
    breakatwhitespace=false,         
    breaklines=true,                 
    captionpos=b,                    
    keepspaces=true,                 
    numbers=left,                    
    numbersep=5pt,                  
    showspaces=false,                
    showstringspaces=false,
    showtabs=false,                  
    tabsize=4,
    frame=tb,
    rulecolor=\color{black!10}
}
\usepackage[textsize=tiny]{todonotes}

\icmltitlerunning{LittleBit-2: Maximizing the Spectral Energy Gain in Sub-1-Bit LLMs via Latent Geometry Alignment}

\begin{document}

\twocolumn[
  \icmltitle{LittleBit-2: Maximizing the Spectral Energy Gain in Sub-1-Bit LLMs\\via Latent Geometry Alignment}
  \icmlsetsymbol{equal}{*}
  \icmlsetsymbol{corres}{$\dagger$}

  \begin{icmlauthorlist}
    \icmlauthor{Banseok Lee}{comp}
    \icmlauthor{Youngmin Kim}{comp}
  \end{icmlauthorlist}

  \icmlaffiliation{comp}{Samsung Research, Seoul, Korea}

  \icmlcorrespondingauthor{Youngmin Kim}{ym1012.kim@samsung.com}
  
  \icmlkeywords{LLM Quantization, Sub-1-bit, Matrix Factorization, Spectral Analysis, ICML}
  
  \vskip 0.3in
]
\printAffiliationsAndNotice{}  

\begin{abstract} 
We identify the Spectral Energy Gain in extreme model compression, where low-rank binary approximations outperform tiny-rank floating-point baselines for heavy-tailed spectra. However, prior attempts fail to realize this potential, trailing state-of-the-art 1-bit methods. We attribute this degradation to Latent Geometry Misalignment: standard singular vectors exhibit high coherence (spiky distribution), the worst-case geometry for binary quantization. To realize this gain, we propose \textbf{LittleBit-2}, a framework employing Internal Latent Rotation and Joint Iterative Quantization (Joint-ITQ). This approach acts as a geometric preconditioner, aligning coherent latent distributions with the binary hypercube with zero inference overhead. Empirically, LittleBit-2 establishes a new state-of-the-art in the sub-1-bit regime (1$\sim$0.1 bpp) on Llama-2 and Llama-3, matching the fidelity of leading 1-bit baselines.
\end{abstract}

\section{Introduction}
\label{sec:introduction}
Scaling Large Language Models (LLMs) drives remarkable performance gains \cite{brown2020language, touvron2023llama} but faces a rigid hardware constraint: the memory wall \cite{gholami2022survey}. A 70B model requires $\sim$140 GB of VRAM in FP16, prohibiting consumer-grade deployment. Consequently, model compression \cite{han2015deep} has shifted from an optimization technique to a deployment necessity. While Post-Training Quantization (PTQ) has standardized 4-bit precision \cite{frantar2022gptq, lin2024awq}, recent works have pushed towards the 1-bit frontier (e.g., BitNet \cite{wang2023bitnet}, OneBit \cite{xu2024onebit}), demonstrating that LLMs can retain capabilities with ternary or binary weights.

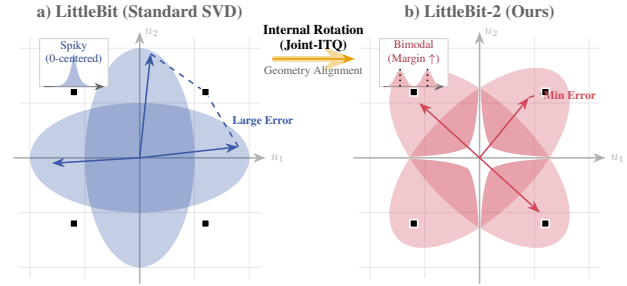
\begin{figure}[t]
\centering
\resizebox{\columnwidth}{!}{%
\begin{tikzpicture}[
    font=\sffamily,
    >=Stealth,
    axis/.style={->, icmlgray!50, thick},
    grid/.style={step=1cm, icmlgray!15, thin},
    binary_node/.style={rectangle, fill=black, minimum size=4pt, inner sep=0pt, draw=white, thick},
    latent_cloud/.style={opacity=0.3, draw=none},
    vector_arrow/.style={->, thick, line cap=round},
    annotation/.style={fill=white, inner sep=2pt, opacity=0.9, text opacity=1, font=\tiny\bfseries, rounded corners=2pt, drop shadow}
]

    \begin{scope}[local bounding box=leftpanel]
        \node[anchor=south, font=\bfseries\small, color=icmlgray] at (0, 2.4) {a) LittleBit (Standard SVD)};
        
        \draw[grid] (-2.2,-2.2) grid (2.2,2.2);
        \draw[axis] (-2.3,0) -- (2.3,0) node[right, scale=0.7] {$u_1$};
        \draw[axis] (0,-2.3) -- (0,2.3) node[right, scale=0.7] {$u_2$};
        
        \node[binary_node] (tr) at (1.2, 1.2) {};
        \node[binary_node] (tl) at (-1.2, 1.2) {};
        \node[binary_node] (bl) at (-1.2, -1.2) {};
        \node[binary_node] (br) at (1.2, -1.2) {};
        
        \fill[icmlblue, latent_cloud] (0,0) ellipse (1.0 and 2.0); 
        \fill[icmlblue, latent_cloud] (0,0) ellipse (2.0 and 1.0); 
        
        \draw[vector_arrow, icmlblue] (0,0) -- (1.8, 0.2);
        \draw[vector_arrow, icmlblue] (0,0) -- (0.2, 1.9);
        \draw[vector_arrow, icmlblue] (0,0) -- (-1.6, -0.1);

        \draw[dashed, icmlblue, thick] (1.8, 0.2) -- (tr) node[midway, right, font=\tiny, xshift=1pt] {\textbf{Large Error}};
        \draw[dashed, icmlblue, thick] (0.2, 1.9) -- (tr);

        \begin{scope}[shift={(-1.2, 1.3)}, scale=0.25]
            \draw[fill=white, draw=icmlgray!30] (-2.5,0) rectangle (2.5, 3.5);
            \draw[->, icmlgray] (-2.2,0) -- (2.2,0);
            \fill[icmlblue!40] plot[domain=-2:2, samples=40] (\x, {2.8*exp(-abs(\x)/0.3)}) -- cycle;
            \node[font=\tiny, align=center, color=icmlblue!80!black] at (0, 2.5) {Spiky\\(0-centered)};
        \end{scope}
    \end{scope}

    \draw[->, line width=3pt, icmlorange!40] ($(leftpanel.east)+(-0.5,1.5)$) -- ($(leftpanel.east)+(1.0,1.5)$);
    \draw[->, thick, icmlorange] ($(leftpanel.east)+(-0.5,1.5)$) -- ($(leftpanel.east)+(1.0,1.5)$)
        node[midway, above, font=\bfseries\scriptsize, align=center, black] {Internal Rotation\\(Joint-ITQ)}
        node[midway, below, font=\tiny, color=icmlgray] {Geometry Alignment};

    \begin{scope}[shift={(6.2,0)}, local bounding box=rightpanel]
        \node[anchor=south, font=\bfseries\small, color=icmlgray] at (0, 2.4) {b) LittleBit-2 (Ours)};
        \draw[grid] (-2.2,-2.2) grid (2.2,2.2);
        \fill[icmlred, latent_cloud, rotate=45] (0,0) ellipse (1.0 and 2.1);
        \fill[icmlred, latent_cloud, rotate=-45] (0,0) ellipse (1.0 and 2.1);
        \fill[white] (-2.3, 0) 
            .. controls (-0.8, 0) and (0, 0.05) .. (0, 0.5) 
            .. controls (0, 0.05) and (0.8, 0) .. (2.3, 0) 
            .. controls (0.8, 0) and (0, -0.05) .. (0, -0.5) 
            .. controls (0, -0.05) and (-0.8, 0) .. cycle;
            
        \fill[white] (0, 2.3) 
            .. controls (0, 0.8) and (0.05, 0) .. (0.5, 0) 
            .. controls (0.05, 0) and (0, -0.8) .. (0, -2.3) 
            .. controls (0, -0.8) and (-0.05, 0) .. (-0.5, 0) 
            .. controls (-0.05, 0) and (0, 0.8) .. cycle;
        
        \draw[axis] (-2.3,0) -- (2.3,0) node[right, scale=0.7] {$u_1$};
        \draw[axis] (0,-2.3) -- (0,2.3) node[right, scale=0.7] {$u_2$};
        
        \node[binary_node] (tr2) at (1.2, 1.2) {}; 
        \node[binary_node] (tl2) at (-1.2, 1.2) {};
        \node[binary_node] (bl2) at (-1.2, -1.2) {};
        \node[binary_node] (br2) at (1.2, -1.2) {};
        
        \draw[vector_arrow, icmlred] (0,0) -- (0.9, 1.1);
        \draw[vector_arrow, icmlred] (0,0) -- (-1.1, 1.0);
        \draw[vector_arrow, icmlred] (0,0) -- (1.1, -1.0);

        \draw[dashed, icmlred, thick] (0.9, 1.1) -- (tr2) node[midway, right, font=\tiny, xshift=1pt] {\textbf{Min Error}};

        \begin{scope}[shift={(-1.2, 1.3)}, scale=0.25]
            \draw[fill=white, draw=icmlgray!30] (-2.5,0) rectangle (2.5, 3.5);
            \draw[->, icmlgray] (-2.2,0) -- (2.2,0);
            \fill[icmlred!40] plot[domain=-2.2:2.2, samples=50] (\x, {1.2*exp(-4*(\x-1)^2) + 1.2*exp(-4*(\x+1)^2)}) -- cycle;
            \node[font=\tiny, align=center, color=icmlred!80!black] at (0, 2.5) {Bimodal\\(Margin $\uparrow$)};
            \draw[dotted, thick] (-1,0) -- (-1,1.5);
            \draw[dotted, thick] (1,0) -- (1,1.5);
        \end{scope}
    \end{scope}

\end{tikzpicture}
}
\vspace{-0.2in}
\caption{\textbf{Latent Geometry Alignment.} 
(a) Standard singular vectors exhibit high coherence (spiky distribution), clustering along the axes. This creates a geometric mismatch with the binary quantization targets (black dots at ($\pm 1$, $\pm 1$)). 
(b) \textbf{LittleBit-2} employs Internal Latent Rotation via Joint-ITQ. Acting as a geometric preconditioner, it rotates latent factors to align with binary hypercube diagonals. This minimizes quantization noise (min error) and maximizes the optimization margin (Bimodal distribution).}
\label{fig:littlebit2_overview}
\vspace{-0.2in}
\end{figure}

However, a critical efficiency gap remains. Even at 1-bit, a 70B model demands $\sim$15 GB of memory, which is still prohibitive for edge devices. This necessitates \textit{sub-1-bit} compression. \textbf{LittleBit} \cite{lee2025littlebit} approached this via a \textbf{Low-Rank Binary} architecture. We validate their design through Spectral Theory \cite{eckart1936approximation}, demonstrating that low-rank binary approximations outperform tiny-rank floating-point counterparts for heavy-tailed weights. Despite this theoretical potential, LittleBit trails the state-of-the-art 1-bit baseline \cite{xu2024onebit}. We attribute their underperformance to Latent Geometry Misalignment.

To realize this gain, we propose \textbf{LittleBit-2}, a framework employing Internal Latent Rotation and Joint Iterative Quantization (Joint-ITQ). As illustrated in Figure \ref{fig:littlebit2_overview}, this acts as a geometric preconditioner, aligning coherent latent distributions with the binary hypercube.

Empirically, LittleBit-2 establishes a new state-of-the-art in the sub-1-bit regime (1$\sim$0.1 bpp) on Llama-2 and Llama-3, matching the fidelity of leading 1-bit baselines.

Our contributions are threefold:
\begin{itemize}
    \item \textbf{Theoretical Diagnosis:} We identify the theoretical superiority of Low-Rank Binary approximation over Tiny-Rank FP16 baselines in heavy-tailed spectra. We formulate this via the Spectral Break-Even Condition, proving that rank expansion outweighs quantization noise.
    \item \textbf{Geometric Alignment via Joint-ITQ:} We propose LittleBit-2, which employs Joint-ITQ to rotate latent factors. This aligns the latent distribution with the binary hypercube vertices, maximizing the decision margin and minimizing quantization error.
    \item \textbf{SOTA Sub-1-bit Performance:} LittleBit-2 establishes a new state-of-the-art in the sub-1-bit regime (down to $\sim$0.1 bpp) on Llama-2 and Llama-3, matching the fidelity of leading 1-bit baselines.
\end{itemize}

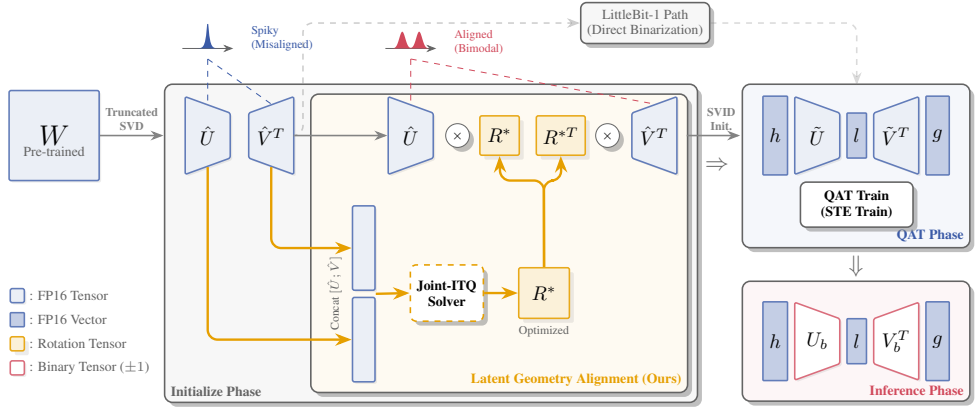
\begin{figure*}[t]
\centering
\resizebox{0.75\textwidth}{!}{%
\begin{tikzpicture}[
    font=\sffamily,
    >=Stealth,
    tensor_fp/.style={
        draw=icmlblue!80, fill=icmlblue!10, thick, rounded corners=1pt,
        drop shadow={opacity=0.15, shadow xshift=1pt, shadow yshift=-1pt}
    },
    tensor_bin/.style={
        draw=icmlred!80, fill=white, pattern color=icmlred!30, thick, rounded corners=1pt,
        drop shadow={opacity=0.15, shadow xshift=1pt, shadow yshift=-1pt}
    },
    tensor_rot/.style={
        draw=icmlorange!90, fill=icmlorange!15, thick, rounded corners=1pt,
        drop shadow={opacity=0.15}
    },
    scale_vec/.style={
        draw=icmlblue!80, fill=icmlblue!30, thick, rounded corners=0.5pt
    },
    process_box/.style={
        rectangle, draw=icmlgray, thick, fill=white, rounded corners=3pt,
        minimum height=0.8cm, align=center, font=\scriptsize\bfseries, drop shadow
    },
    itq_box/.style={
        rectangle, draw=icmlorange, thick, fill=white, dashed, rounded corners=3pt,
        minimum height=1.0cm, align=center, font=\scriptsize\bfseries, drop shadow
    },
    op_circle/.style={
        circle, fill=white, draw=icmlgray, inner sep=0.5pt, minimum size=12pt, font=\tiny, drop shadow
    },
    lora_base_fp/.style={
        trapezium, 
        draw=icmlblue!80, thick, 
        fill=icmlblue!10,
        rounded corners=1pt,
        align=center, 
        font=\small,
        drop shadow={opacity=0.15, shadow xshift=1pt, shadow yshift=-1pt}
    },
    lora_down_fp/.style={
        lora_base_fp,
        trapezium angle=70,      
        shape border rotate=270,  
        minimum height=0.8cm     
    },
    lora_up_fp/.style={
        lora_base_fp,
        trapezium angle=70,
        shape border rotate=90, 
        minimum height=0.8cm
    },
    lora_base_binary/.style={
        trapezium, 
        draw=icmlred!80, fill=white, pattern color=icmlred!30, thick, 
        rounded corners=1pt,
        align=center, 
        font=\small,
        drop shadow={opacity=0.15, shadow xshift=1pt, shadow yshift=-1pt}
    },
    lora_down_binary/.style={
        lora_base_binary,
        trapezium angle=70,      
        shape border rotate=270,  
        minimum height=0.8cm     
    },
    lora_up_binary/.style={
        lora_base_binary,
        trapezium angle=70,
        shape border rotate=90, 
        minimum height=0.8cm
    },
    arrow_line/.style={->, thick, color=icmlgray!80, rounded corners=5pt},
    main_arrow/.style={->, very thick, color=icmlorange, rounded corners=5pt},
    label_txt/.style={font=\scriptsize, color=icmlgray}
]
    \node[tensor_fp, minimum height=1.6cm, minimum width=1.6cm] (W) at (0,0) {};
    \node at (W.center) {\Large $W$};
    \node[below=2pt, label_txt] at (W.center) {Pre-trained};
    \draw[arrow_line] (W.east) -- node[above, midway, font=\bfseries\tiny, align=center, yshift=-1pt] {Truncated\\SVD} ([xshift=1.13cm]W.east);
    \node[lora_down_fp, right=1.5cm of W] (U_init) {};
    \node at (U_init.center) {$\hat{U}$};
    \node[lora_up_fp, right=0.3cm of U_init] (Vt_init) {};
    \node at (Vt_init.center) {$\hat{V}^T$};
    \node[above=0.8cm of U_init] (hist_spiky) {
        \begin{tikzpicture}[scale=0.15, baseline]
            \draw[->, icmlgray] (-3,0) -- (3,0);
            \fill[icmlblue] plot[domain=-2.5:2.5, samples=40] (\x, {4*exp(-abs(\x)/0.2)}) -- cycle;
        \end{tikzpicture}
    };
    \node[right=2pt of hist_spiky, font=\tiny, align=left, color=icmlblue] {Spiky\\(Misaligned)};
    \draw[dashed, icmlblue, thin] (hist_spiky.south) -- (U_init.north);
    \draw[dashed, icmlblue, thin] (hist_spiky.south) -- (Vt_init.north);
    \matrix [
        anchor=north west,          
        row sep=0.15cm,             
        column sep=0.08cm,           
        inner sep=0cm,            
        nodes={anchor=center},
        ampersand replacement=\&
    ] at (-0.8, -2.7) {             
        \node[tensor_fp, minimum size=0.25cm](legend_tfp) {}; \& 
        \node[label_txt, anchor=west, right=0cm of legend_tfp] {: FP16 Tensor}; \\
        
        \node[scale_vec, minimum height=0.25cm, minimum width=0.25cm](legend_sv) {}; \& 
        \node[label_txt, anchor=west, right=0cm of legend_sv] {: FP16 Vector}; \\
        
        \node[tensor_rot, minimum size=0.25cm](legend_tr) {}; \& 
        \node[label_txt, anchor=west, right=0cm of legend_tr] {: Rotation Tensor}; \\
        
        \node[tensor_bin, minimum size=0.25cm](legend_tb) {}; \& 
        \node[label_txt, anchor=west, right=0cm of legend_tb] {: Binary Tensor ($\pm 1$)}; \\
    };
    \node[process_box, minimum height=0.6cm, font=\scriptsize, fill=icmlgray!5, text=icmlgray] (lb1_box) at (10.5, 2.0) {LittleBit-1 Path\\(Direct Binarization)};
    \draw[arrow_line, dashed, opacity=0.4] (Vt_init.east) -- ++(0.15,0) |- (lb1_box.west);
    \node[right=1.8cm of Vt_init, yshift=-1.5cm] (stack_center) {};
    \node[tensor_fp, minimum height=1.5cm, minimum width=0.4cm] (V_stack) at (5.5, -2.0) {};
    \node[tensor_fp, minimum height=1.5cm, minimum width=0.4cm, below=0.1cm of V_stack] (U_stack) {};
    \node[rotate=90, font=\tiny, color=icmlgray] (concat) at ($(V_stack.west)+(-0.3,-0.8)$) {Concat $[\hat{U};\hat{V}]$};
    \draw[main_arrow] (U_init.south) -- ++(0,-0.5) |- (U_stack.west);
    \draw[main_arrow] (Vt_init.south) -- ++(0,-0.5) |- (V_stack.west);
    \node[itq_box, right=0.6cm of V_stack, yshift=-0.8cm] (itq_solver) {
        \textbf{Joint-ITQ}\\Solver
    };
    \draw[main_arrow] ($(V_stack.east)!0.5!(U_stack.east)$) -- (itq_solver.west);
    \node[tensor_rot, right=0.6cm of itq_solver, minimum size=0.9cm] (R_mat) {$R^*$};
    \draw[main_arrow] (itq_solver.east) -- (R_mat.west);
    \node[below, font=\tiny, color=icmlgray] at (R_mat.south) {Optimized};

    \node[lora_down_fp, right=1.65cm of Vt_init] (U_rot) {};
    \node[lora_up_fp, right=3.5cm of U_rot, anchor=west] (V_rot) {};
    \node[op_circle, right=0.2cm of U_rot] (mult_u) {$\times$};
    \node[op_circle, left=0.2cm of V_rot] (mult_v) {$\times$};
    \node[tensor_rot, right=0.8cm of U_rot, minimum size=0.7cm] (Left_R_mat) {$R^*$};
    \node[tensor_rot, left=0.8cm of V_rot, minimum size=0.7cm] (Right_R_mat) {$R^{*T}$};
    \node at (U_rot.center) {$\hat{U}$};
    \node at (V_rot.center) {$\hat{V}^T$};
    \draw[arrow_line] (Vt_init.east) -- (U_rot.west);
    \draw[main_arrow] (R_mat.north) -- (8.7,-0.8) -- (7.925,-0.8) -- (Left_R_mat.south);
    \draw[main_arrow] (R_mat.north) -- (8.7,-0.8) -- (9.01,-0.8) -- (Right_R_mat.south);
    \node[above=0.8cm of U_rot] (hist_aligned) {
        \begin{tikzpicture}[scale=0.15, baseline]
            \draw[->, icmlgray] (-3,0) -- (3,0);
            \fill[icmlred] plot[domain=-2.5:2.5, samples=50] (\x, {2*exp(-4*(\x-1)^2) + 2*exp(-4*(\x+1)^2)}) -- cycle;
        \end{tikzpicture}
    };
    \node[right=2pt of hist_aligned, font=\tiny, align=left, color=icmlred] {Aligned\\(Bimodal)};
    \draw[dashed, icmlred, thin] (hist_aligned.south) -- (U_rot.north);
    \draw[dashed, icmlred, thin] (hist_aligned.south) -- (V_rot.north);

    \node[scale_vec, minimum height=1.4cm, minimum width=0.2cm, right=1.5cm of V_rot] (h_fp) {$h$};
    \node[lora_down_fp, right=0.1cm of h_fp] (U_fp) {};
    \node at (U_fp.center) {$\tilde{U}$};
    \node[scale_vec, minimum height=0.8cm, minimum width=0.2cm, right=0.1cm of U_fp] (l_fp) {$l$};
    \node[lora_up_fp, right=0.1cm of l_fp] (Vt_fp) {};
    \node at (Vt_fp.center) {$\tilde{V}^T$};
    \node[scale_vec, minimum height=1.4cm, minimum width=0.2cm, right=0.1cm of Vt_fp] (g_fp) {$g$};
    \draw[arrow_line] ([xshift=0.16cm]V_rot.east) -- node[above, midway, font=\bfseries\tiny, align=center, yshift=-1pt, xshift=4pt] {SVID\\Init.} ([xshift=1.11cm]V_rot.east);
    \node[process_box, minimum width=2.0cm, below=0.4cm of l_fp] (svid_box) {QAT Train\\ (STE Train)}; 

    \draw[arrow_line, dashed, opacity=0.3] (lb1_box.east) -| ([yshift=0.48cm]l_fp.north);
    
    \node[scale_vec, minimum height=0.8cm, minimum width=0.2cm, below=1.6cm of svid_box ] (l_fin) {$l$};
    \node[lora_down_binary, left=0.1cm of l_fin] (U_fin) {};
    \node at (U_fin.center) {$U_b$};
    \node[scale_vec, minimum height=1.4cm, minimum width=0.2cm, left=0.1cm of U_fin] (h_fin) {$h$};
    \node[lora_up_binary, right=0.1cm of l_fin] (Vt_fin) {};
    \node at (Vt_fin.center) {$V_b^T$};
    \node[scale_vec, minimum height=1.4cm, minimum width=0.2cm, right=0.1cm of Vt_fin] (g_fin) {$g$};
    
    \begin{scope}[on background layer]
        \node[process_box, fit=(U_init)(U_stack)(itq_solver)(R_mat)(U_rot)(V_rot), fill=icmlgray!5, rounded corners, inner sep=10pt] (highlight_bg_init) {};
        \node[anchor=south west, font=\bfseries\scriptsize, color=icmlgray] at (highlight_bg_init.south west) {Initialize Phase};
        
        \node[process_box, fit=(svid_box)(U_fp)(Vt_fp)(h_fp.west)(g_fp.east), fill=icmlblue!5, rounded corners, inner sep=10pt] (highlight_bg_train) {};
        \node[anchor=south east, font=\bfseries\scriptsize, color=icmlblue] at (highlight_bg_train.south east) {QAT Phase};
        
        \node[left=0.1cm of highlight_bg_train, font=\large\bfseries, color=icmlgray] (eq) {$\Rightarrow$};
        
        \node[process_box, fit=(g_fin)(h_fin), fill=icmlred!5, rounded corners, inner sep=10pt] (highlight_bg_inference) {};
        \node[anchor=south east, font=\bfseries\scriptsize, color=icmlred] at (highlight_bg_inference.south east) {Inference Phase};
        
        \node[above=-0.05cm of highlight_bg_inference, font=\large\bfseries, color=icmlgray] (eq) {$\Downarrow$};
        
        \node[process_box, fit=(U_stack)(itq_solver)(R_mat)(U_rot)(V_rot)(concat), fill=icmlorange!5, rounded corners, inner sep=5pt] (highlight_bg) {};
        \node[anchor=south east, font=\bfseries\scriptsize, color=icmlorange] at (highlight_bg.south east) {Latent Geometry Alignment (Ours)};
    \end{scope}
\end{tikzpicture}
}
\caption{\textbf{The LittleBit-2 Framework Pipeline.} 
Starting from a truncated SVD of the pretrained weight $W$, \textbf{LittleBit-2} (lower path) explicitly addresses the geometric misalignment. 
The factors $\hat{U}$ and $\hat{V}$ are concatenated and fed into the \textbf{Joint-ITQ} solver to optimize an orthogonal rotation $R$. 
This rotation is applied ($\times$) to $\hat{U}$ and $\hat{V}$, transforming the spiky latent distribution (blue histogram) into an aligned bimodal distribution (red histogram). 
Finally, Dual-SVID and QAT extract the FP16 scales ($h, l, g$) and learn the binary factors ($U_b, V_b^T$).}
\label{fig:littlebit2_framework}
\vspace{-0.2in}
\end{figure*}

\section{Related Work}
\label{sec:related_work}
\subsection{Extreme Quantization}
Recent approaches to 1-bit LLMs, such as BitNet b1.58 \cite{wang2023bitnet, ma2024era} introduced ternary weights $\{-1, 0, +1\}$, demonstrating that LLMs can be trained from scratch with extreme quantization. However, these require prohibitive training cost compared to compressing pre-trained weights. Alternatively, several PTQ frameworks (e.g., BiLLM \cite{huang2024billm}, ARB-LLM \cite{li2024arb}, STBLLM \cite{dong2024stbllm}) achieve nominal sub-1-bit precision by retaining salient weights in high precision while binarizing the rest. However, these methods rely on auxiliary binary masks and sparse indices to locate outliers. In practice, this metadata introduces additional memory overhead. 

1-bit and sub-1-bit QAT presents unique challenges. Unlike moderate compression where PTQ suffices, this regime hits an irreducible error floor due to the collapse of representational capacity. Consequently, \textbf{Quantization-Aware Knowledge Distillation (QAKD)} \cite{kim2019qkd, liu2024llm, du2024bitdistiller} has emerged as a highly effective strategy to recover fidelity. Existing methods like OneBit \cite{xu2024onebit} and LittleBit \cite{lee2025littlebit} utilize this paradigm to map weights to extremely low-bit domains. \textbf{Building on this protocol, our work specifically tackles the initialization bottleneck via geometric alignment.}

\subsection{Incoherence Processing}
To mitigate quantization errors caused by weight outliers, Incoherence Processing via orthogonal rotation has emerged as a critical preconditioner. Existing methods like QuIP\# \cite{tseng2024quip}, QuaRot \cite{ashkboos2024quarot}, and SpinQuant \cite{liu2024spinquant} apply global rotations, specifically learned or randomized Hadamard transforms, to full weight and activation matrices. While this effectively delocalizes outliers into quantization-friendly Gaussian distributions, it often incurs inference overhead due to the requisite online Hadamard transforms. In contrast, LittleBit-2 applies rotation strictly to the latent factors ($\hat{U}, \hat{V}$) during the weight decomposition process. This approach isolates geometric alignment to the initialization phase, effectively resolving the optimization instability \textbf{without incurring any additional inference overhead compared to LittleBit}.

\section{Background}
\label{sec:background}
\subsection{The LittleBit Architecture}
\label{subsec:littlebit_arch}
Conventional quantization approximates the weight matrix $W \in \mathbb{R}^{d_{out} \times d_{in}}$, facing a lower bound of 1 bit per parameter (bpp) even with binary weights. To achieve sub-1-bit compression, LittleBit \cite{lee2025littlebit} utilizes a Latent Factorization Quantization framework. It decouples the parameter count from matrix dimensions via low-rank factorization ($U_r$, $V_r$), enabling fractional effective bit-rates (e.g., $< 1$ bpp) even with binary weights.

\paragraph{Tri-Scale Latent Factorization.}
The framework decomposes $W$ into binary latent factors $U_{b} \in \{\pm 1\}^{d_{out} \times r}$ and $V_{b} \in \{\pm 1\}^{d_{in} \times r}$ with rank $r \ll \min(d_{in}, d_{out})$. To explicitly decouple magnitude from sign (i.e., $U \approx |U| \odot U_{b}$), LittleBit approximates the magnitude envelopes $|U|$ and $|V|$ via Rank-1 factorization. Consequently, LittleBit employs a symmetric \textbf{Scale-Binary-Scale-Binary-Scale} architecture to recover magnitude precision. It defines three learnable FP16 scaling components: row scale $h \in \mathbb{R}^{d_{out}}$, a central latent scale vector $l \in \mathbb{R}^{r}$, and column scale $g \in \mathbb{R}^{d_{in}}$, formulated as:
\begin{equation}
    \hat{W} = \underbrace{\text{diag}(h)}_{\text{Scale}} \cdot \underbrace{U_{b}}_{\text{Binary}} \cdot \underbrace{\text{diag}(l)}_{\text{Scale}} \cdot \underbrace{V_{b}^T}_{\text{Binary}} \cdot \underbrace{\text{diag}(g)}_{\text{Scale}}
    \label{eq:factorization}
\end{equation}
This structure ensures that the binary latent factors are sandwiched by floating-point scales, allowing for extreme compression ratios by controlling the rank $r$ while preserving the magnitude dynamics of the original weights. Following LittleBit, we employ this in a residual structure repeated twice; see Appendix \ref{app:residual} for details.

\paragraph{Initialization.}
Training latent factors ($\hat{U}$, $\hat{V}$) from scratch is unstable due to the sign function's gradient mismatch. LittleBit mitigates this via a Dual-SVID strategy, initializing latent factors from singular vectors (binarized as $U_b = \text{sign}(U)$ during the forward pass \cite{bengio2013estimating}). To derive the FP16 scales ($h, l, g$), it performs Rank-1 approximation on the absolute values of these vectors (i.e., $\text{SVD}_1(|\hat{U}|)$) to explicitly decouple the magnitude envelope from the sign (Appendix \ref{app:littlebit_details}). \textbf{In this work, we do not propose a new architecture but refine the initialization to stabilize training and boost performance, ensuring robust optimization.}

\subsection{Latent Geometry Misalignment}
\label{subsec:gap}
While LittleBit achieves sub-1-bit compression, a performance gap remains compared to state-of-the-art 1-bit models. For instance, under similar memory budgets, \textbf{OneBit} \cite{xu2024onebit} achieves a perplexity of 8.36 on Llama-2-7B, whereas LittleBit lags at 9.08.

We attribute this performance degradation to a \textbf{Latent Geometry Misalignment}. The latent factors derived from standard SVD frequently exhibit high coherence, inheriting the spiky nature of the weight distribution \cite{tseng2024quip, liu2024spinquant} in a small subset of dominant channels. In these dimensions, information concentrates in a few outliers while the majority of values remain near zero. This represents a worst-case scenario for binary quantization ($\text{sign}(x)$), as binarization destroys the latent structure, causing high quantization noise. \textbf{LittleBit-2 aims to resolve this by geometrically realigning the latent factors before binarization to maximize information retention}.

\section{The LittleBit-2 Framework}
\label{sec:method}
In this section, we formulate the sub-1-bit quantization as a constrained spectral optimization problem. We derive the Spectral Break-Even Condition, establishing the condition where the information gain from rank expansion outweighs the quantization noise. We then propose \textbf{LittleBit-2}, which satisfies this condition via Latent Geometry Alignment, solving a joint orthogonal Procrustes problem to maximize the geometric margin of quantized factors.

\subsection{Why Low-Rank Binary?}
\label{sec:why_binary}
\paragraph{Problem Formulation}
We formulate sub-1-bit quantization as a constrained optimization problem to justify Low-Rank Binary architectures \cite{lee2025littlebit}. Given a pre-trained weight $W \in \mathbb{R}^{d_{out} \times d_{in}}$ with rank $d = \min(d_{in}, d_{out})$, we seek an approximation $\hat{W}$ minimizing reconstruction error $\|\cdot\|_F^2$ ($\coloneq \mathcal{E}$) under a strict bit-budget $\mathcal{B}$:
\begin{equation}
    \min_{\hat{W}} \| W - \hat{W} \|_F^2 \quad \text{s.t.} \quad \text{Bits}(\hat{W}) \le \mathcal{B}
    \label{eq:opt_problem}
\end{equation}
We analyze two distinct strategies under the budget $\mathcal{B}$:
\begin{itemize}
    \item \textbf{Strategy A (Tiny-Rank FP16):} Retains minimal rank $r_A$. Error is dominated by truncation: $\mathcal{E}_A \approx \mathcal{E}_{\text{trunc}}(r_A)$.
    \item \textbf{Strategy B (Low-Rank Binary):} Retains expanded rank $r_B \approx 16 r_A$ with 1-bit quantization. The error comprises reduced truncation and quantization noise: $\mathcal{E}_B \approx \mathcal{E}_{\text{trunc}}(r_B) + \mathcal{E}_{\text{quant}}(r_B)$, aligning with LittleBit.
\end{itemize}
\paragraph{Spectral Analysis} We analyze the trade-off under a fixed budget $\mathcal{B}$, modeling singular values with a power-law decay $\sigma_k \approx C k^{-\gamma}$ characteristic of LLM weights~\cite{martin2021implicit}. Following their definition, we classify the spectrum as heavy-tailed if $\gamma \leq 0.5$ and light-tailed if $\gamma > 0.5$. For analytical tractability, we approximate the discrete spectrum via continuous integration.

\begin{proposition}[Spectral Break-Even Condition]
\label{thm:superiority}
Let the quantization noise be $\mathcal{E}_{\text{quant}}(r) = \int_{0}^{r} \Lambda \sigma(x)^2dx$ and truncation error be $\mathcal{E}_{\text{trunc}}(r) = \int_{r}^{\infty} \sigma(x)^2dx$, assuming an average distortion coefficient $\Lambda$.
Strategy B (binary, rank $r_B$) outperforms Strategy A (FP16, rank $r_A \ll r_B$) if and only if the \textbf{Tail Energy Gain} outweighs the \textbf{Quantization Cost}:
\begin{equation}
    \underbrace{\int_{r_A}^{r_B} \sigma(x)^2dx}_{\text{Tail Gain}} > \underbrace{\int_{0}^{r_B} \Lambda\sigma(x)^2 dx}_{\text{Quantization Cost}}
    \label{eq:breakeven}
\end{equation}
This implies a critical threshold $\gamma^*$; Strategy B is superior for heavy-tailed distributions ($\gamma < \gamma^*$).
\end{proposition}
Proposition \ref{thm:superiority} establishes a critical spectral decay rate $\gamma^*$ such that for any distribution with $\gamma < \gamma^*$, the Low-Rank Binary strategy is theoretically superior to the Tiny-Rank FP16 approximation. This trade-off is visualized in Figure~\ref{fig:spectral_gap}, provided in Appendix~\ref{app:spectral_gap_viz}.
Empirically, Llama-2 7B exhibits a break-even point at $\gamma^* \approx 0.36$ (Section \ref{subsec:synthetic_validation}). 73\% of linear layers exhibit $\gamma < 0.36$, confirming that the heavy-tailed structure of modern LLM weights intrinsically favors the Low-Rank Binary architecture over FP16 truncation.

As $\gamma$ is intrinsic to pre-trained weights, the distortion coefficient $\Lambda$ remains the only controllable variable. Minimizing $\Lambda$ shifts the break-even threshold $\gamma^*$ higher, extending applicability to lighter-tailed models, and widens the margin between Tail Gain and Quantization Cost (Eq. \ref{eq:breakeven}), reducing approximation error.

\subsection{The Distortion Coefficient $\Lambda$}
While treated as a constant in Proposition \ref{thm:superiority}, $\Lambda$ is practically governed by latent factor geometry. With $\gamma$ fixed, we focus on minimizing $\Lambda$ by analyzing the relationship between vector geometry and quantization noise.

Consider a latent vector $\mathbf{u} \in \mathbb{R}^r$ (a row of latent factor $\hat{U}$). In our low-rank binary architecture, $\mathbf{u}$ is approximated by a binary vector $\mathbf{b} = \text{sign}(\mathbf{u})$ and an optimal scalar scale $h_i \in \mathbb{R}$. The quantization error is defined as $\mathcal{E}(\mathbf{u}) = \min_\alpha \| \mathbf{u} - \alpha \mathbf{b} \|_2^2$.\footnote{While LittleBit employs Rank-1 approximation ($|U| \approx h l_u^T$) in practice, we analyze the channel-wise scale case here for analytical tractability. This abstraction is valid because scalar distortion serves as a theoretical lower bound for Rank-1 error; the geometric dependency remains the fundamental determinant.}

We define the \textbf{Local Distortion Coefficient} as the ratio of the error energy to the signal energy: $\lambda(\mathbf{u}) \coloneq \mathcal{E}(\mathbf{u}) / \| \mathbf{u} \|_2^2$.
\begin{lemma}[Distortion-Geometry Duality]
\label{lemma:distortion_duality}
The optimal local distortion coefficient is strictly determined by the ratio of the squared $L_1$ norm to the squared $L_2$ norm of the vector:
\begin{equation}
    \lambda(\mathbf{u}) = 1 - \frac{1}{r} \left( \frac{\| \mathbf{u} \|_1}{\| \mathbf{u} \|_2} \right)^2
    \label{eq:lambda_proof}
\end{equation}
\end{lemma}
Equation \ref{eq:lambda_proof} (proven in Appendix \ref{app:proof_distortion}) shows that noise minimization is equivalent to maximizing the vector denseness ($\| \mathbf{u} \|_1 / \| \mathbf{u} \|_2$). This ratio is inversely related to the Coordinate Incoherence defined in the next section, serving as the bridge between geometry and quantization fidelity. Since the weight is reconstructed via $\hat{W} \approx \hat{U}\hat{V}^T$, assuming statistical independence between the quantization errors of the factors, the global distortion $\Lambda$ compounds the local distortions of both interacting factors:
\begin{equation}
    \Lambda_{i,j} \approx 1 - (1 - \lambda_u(\mathbf{u}_i))(1 - \lambda_v(\mathbf{v}_j))
\end{equation}
Since minimizing the global distortion $\Lambda$ essentially reduces to suppressing the local distortion $\lambda$ of the factors, we focus our subsequent analysis on the local geometry defined by $\lambda$.

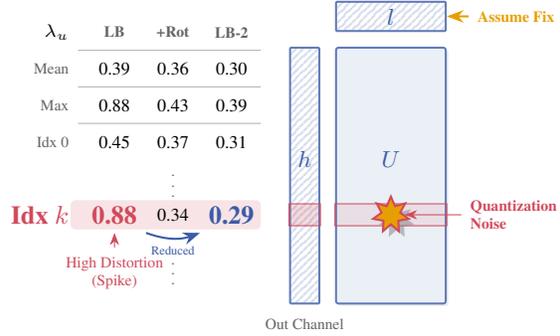
\begin{figure}[t]
\centering
\resizebox{!}{4.5cm}{%
\begin{tikzpicture}[
    font=\sffamily,
    >=Stealth,
    tensor_blue/.style={
        draw=icmlblue!80, fill=icmlblue!10, thick, rounded corners=1pt,
        drop shadow={opacity=0.15, shadow xshift=1pt, shadow yshift=-1pt}
    },
    tensor_gray/.style={
        draw=icmlgray!80, fill=icmlgray!10, thick, rounded corners=1pt
    },
    scale_box/.style={
        draw=icmlorange!90, fill=icmlorange!10, thick, rounded corners=0.5pt, 
        font=\tiny\bfseries, align=center
    },
    scale_vec/.style={
        draw=icmlblue!80, fill=icmlblue!5, thick, rounded corners=0.5pt,
        pattern={Lines[distance=2pt, angle=45]}, pattern color=icmlblue!20
    },
    highlight_row/.style={
        fill=icmlred!15, draw=none
    },
    noise_flash/.style={
        star, star points=7, star point height=0.12cm, minimum size=0.5cm, 
        fill=icmlorange, draw=icmlred, thick, drop shadow, inner sep=0pt
    },
    table_txt/.style={font=\scriptsize, anchor=center}
]

    \node[tensor_blue, minimum height=3.5cm, minimum width=1.5cm] (U) at (0,0) {};
    \node[above=0pt, font=\bfseries\small, color=icmlblue] at (U) {$U$};

    \node[scale_vec, minimum height=3.5cm, minimum width=0.4cm, left=0.2cm of U] (h) {};
    \node[above=0pt, font=\bfseries\small, color=icmlblue] at (h.center) {$h$};
    \node[below=2pt, font=\tiny, color=icmlgray] at (h.south) {Out Channel};

    \node[scale_vec, minimum height=0.4cm, minimum width=1.5cm, above=0.2cm of U.north] (l) {};
    \node[font=\bfseries\small, color=icmlblue] at (l.center) {$l$};

    \node[right=0.3cm of l, font=\tiny\bfseries, color=icmlorange, align=center] (fix_lbl) {Assume Fix};
    \draw[->, thick, icmlorange] (fix_lbl.west) -- (l.east);

    \coordinate (row_start) at ($(h.north) + (-1.7, -0.3)$);
    \def\rowstep{0.5}
    
    \coordinate (r1) at (row_start);
    \coordinate (r2) at ($(r1) + (0, -\rowstep)$);
    \coordinate (r3) at ($(r2) + (0, -\rowstep)$);
    \coordinate (r4) at ($(r3) + (0, -\rowstep)$); 
    \coordinate (r5) at ($(r4) + (0, -\rowstep)$); 
    \coordinate (r6) at ($(r5) + (0, -\rowstep - 0.2)$); 
    \coordinate (r7) at ($(r6) + (0, -\rowstep - 0.2)$); 

    \def\colLB{-0.9}
    \def\colRot{-0.1}
    \def\colLBTWO{0.7}

    \node[font=\tiny\bfseries, color=icmlgray] at ($(r1) + (\colLB, 0.5)$) {LB};
    \node[font=\tiny\bfseries, color=icmlgray] at ($(r1) + (\colRot, 0.5)$) {+Rot};
    \node[font=\tiny\bfseries, color=icmlgray] at ($(r1) + (\colLBTWO, 0.5)$) {LB-2};
    \node[font=\scriptsize\bfseries, color=black, anchor=east] at ($(r1) + (-1.4, 0.5)$) {$\lambda_u$};
    \draw[icmlgray!50] ($(r1) + (-1.4, 0.25)$) -- ($(r1) + (1.1, 0.25)$);
    \draw[icmlgray!50] ($(r2) + (-1.4, 0.25)$) -- ($(r2) + (1.1, 0.25)$);
    \draw[icmlgray!50] ($(r3) + (-1.4, 0.25)$) -- ($(r3) + (1.1, 0.25)$);
    \node[table_txt] at ($(r1) + (\colLB, 0)$) {0.39};
    \node[table_txt] at ($(r1) + (\colRot, 0)$) {0.36};
    \node[table_txt] at ($(r1) + (\colLBTWO, 0)$) {0.30};
    \node[anchor=east, font=\tiny, color=icmlgray] at ($(r1) + (-1.4, 0)$) {Mean};

    \node[table_txt] at ($(r2) + (\colLB, 0)$) {0.88};
    \node[table_txt] at ($(r2) + (\colRot, 0)$) {0.43};
    \node[table_txt] at ($(r2) + (\colLBTWO, 0)$) {0.39};
    \node[anchor=east, font=\tiny, color=icmlgray] at ($(r2) + (-1.4, 0)$) {Max};

    \node[table_txt] at ($(r3) + (\colLB, 0)$) {0.45};
    \node[table_txt] at ($(r3) + (\colRot, 0)$) {0.37};
    \node[table_txt] at ($(r3) + (\colLBTWO, 0)$) {0.31};
    \node[anchor=east, font=\tiny, color=icmlgray] at ($(r3) + (-1.4, 0)$) {Idx 0};

    \node[font=\tiny, color=icmlgray] at ($(r4) + (\colRot, 0)$) {$\vdots$};

    \draw[highlight_row, rounded corners=2pt] ($(r5) + (-1.5, 0.2)$) rectangle ($(r5) + (1.1, -0.2)$);
    
    \node[table_txt, color=icmlred, font=\bfseries] (val_bad) at ($(r5) + (\colLB, 0)$) {0.88};
    \node[table_txt] at ($(r5) + (\colRot, 0)$) {0.34};
    \node[table_txt, color=icmlblue, font=\bfseries] (val_good) at ($(r5) + (\colLBTWO, 0)$) {0.29};
    \node[anchor=east, font=\tiny, color=icmlred, font=\bfseries] at ($(r5) + (-1.4, 0)$) {Idx $k$};

    \node[font=\tiny, color=icmlgray] at ($(r6) + (\colRot, 0)$) {$\vdots$};

    \path let \p1 = (r5) in coordinate (y_bad) at (0, \y1);
    
    \draw[fill=icmlred, fill opacity=0.2, draw=icmlred] 
        ($(h.west |- y_bad) + (0, 0.15)$) rectangle ($(h.east |- y_bad) + (0, -0.15)$);
        
    \draw[fill=icmlred, fill opacity=0.1, draw=icmlred] 
        ($(U.west |- y_bad) + (0, 0.15)$) rectangle ($(U.east |- y_bad) + (0, -0.15)$);
    \node[noise_flash] (flash) at ($(U.center |- y_bad)$) {};
    
    \node[anchor=west, font=\tiny\bfseries, color=icmlred, align=left, xshift=0.2cm] 
        (noise_txt) at (U.east |- y_bad) {Quantization\\Noise};
        
    \draw[->, icmlred] (noise_txt.west) -- (flash.east);
    \node[below=0.2cm of val_bad, font=\tiny, color=icmlred, align=center] (spike_txt) {High Distortion\\(Spike)};
    \draw[->, icmlred] (spike_txt.north) -- (val_bad.south);
    \draw[->, icmlblue, thick, bend right=20] (val_bad.south east) to node[below, font=\tiny, scale=0.8] {Reduced} (val_good.south west);
\end{tikzpicture}
}
\caption{\textbf{Latent Geometry Misalignment.} We visualize the local distortion coefficient $\lambda$ across latent rows. Standard initialization (LB) suffers from geometric outliers (max $\lambda \approx 0.88$), which act as spikes that degrade the precision of the shared floating-point scales. LittleBit-2 effectively suppresses these outliers through internal rotation, reducing the peak distortion to $0.29$ and minimizing quantization noise.}
\label{fig:geometry_misalignment}
\vspace{-0.2in}
\end{figure}

Eq. \ref{eq:breakeven} requires minimizing $\lambda$. However, standard SVD typically yields latent factors with near-worst-case $\lambda$, exhibiting high coherence in dominant channels. This concentration creates a high-distortion barrier, violating the inequality in Eq. \ref{eq:breakeven}.

\begin{definition}[Coordinate Incoherence]
\label{def:coherence}
Let $U \in \mathbb{R}^{d_{out} \times r}$ be an orthogonal matrix. We define the coordinate incoherence $\mu(U)$ based on the element-wise infinity norm:
\begin{equation}
    \mu(U) = \sqrt{d} \max_{i,j} |U_{ij}|
\end{equation}
High $\mu$ implies energy concentration (spiky), whereas low $\mu$ indicates uniform information spread.
\end{definition}

To quantify the impact of coherence on quantization, we contrast two geometric extremes derived from Lemma \ref{lemma:distortion_duality}. The distortion $\lambda$ is strictly governed by the vector's density, which links directly to the Coordinate Coherence $\mu(U)$. In the worst-case scenario (High Coherence), latent vectors align with coordinate axes (e.g., $\mathbf{u} \approx [1, 0, \dots, 0]$). In this sparse regime, the $L_1$ norm collapses to the $L_2$ norm, pushing the distortion to its theoretical maximum $\lambda_{\text{worst}} \approx 1.0$, where binary quantization fails completely. Conversely, the best-case scenario (Low Coherence) occurs when the vector energy is democratized across all dimensions (dense). Here, the ratio $\| \mathbf{u} \|_1 / \| \mathbf{u} \|_2$ is maximized, driving the distortion $\lambda$ toward zero. As illustrated in Figure \ref{fig:geometry_misalignment}, standard initialization suffers from these high-distortion spikes, whereas our geometric alignment successfully suppresses them.

\begin{figure}[t]
    \centering
    \includegraphics[width=1.0\linewidth]{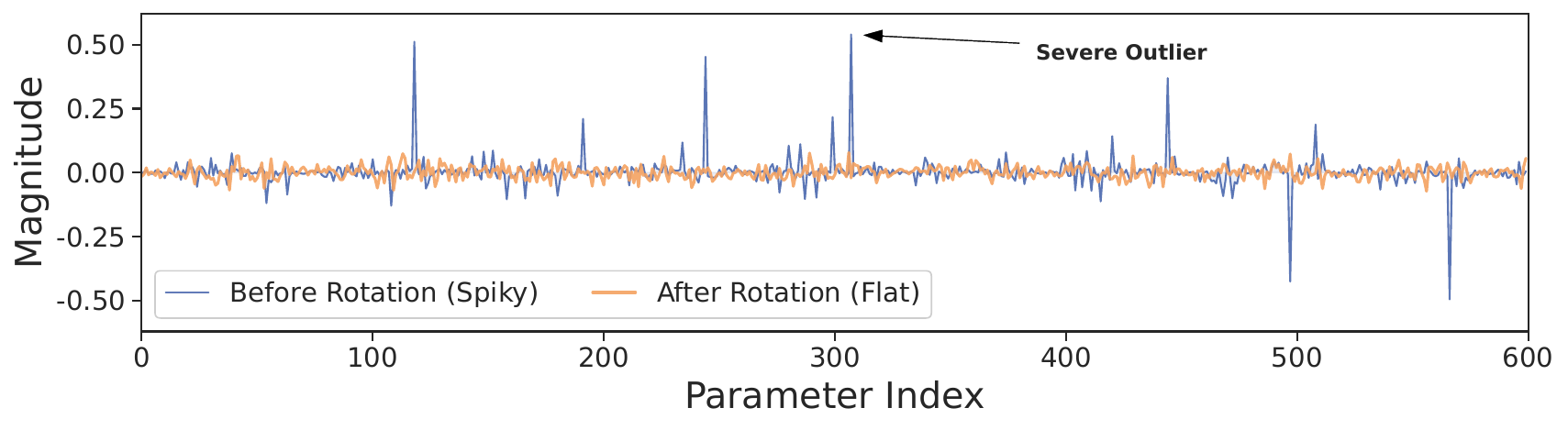}
    \vspace{-0.2in}
    \caption{\textbf{Visualization of Latent Geometry Alignment.} 
    Histograms of latent factor $\hat{U}$ derived from the \textit{q\_proj} ($W_Q$) in the middle (15th) layer of Llama-2 7B. Applying Internal Rotation transforms the distribution into a Gaussian (Orange), effectively mitigating outliers.
    }
    \label{fig:histogram_alignment}
    \vspace{-0.2in}
\end{figure}

Analyzing the latent factors of Llama-2 7B (15th layer \texttt{q\_proj}, Figure \ref{fig:histogram_alignment}), we observe that standard singular vectors exhibit a highly skewed distribution (Kurtosis $\approx 16.8$) with heavy tails reaching maximum distortions of $\lambda_{\text{max}} \approx 0.88$. This empirical proximity to the worst-case barrier demonstrates that standard singular vectors inherently possess a geometry hostile to binary quantization.

\subsection{Latent Geometry Preconditioning}
\label{subsec:internal_incoherence_processing}
To mitigate \textit{High-Distortion Barrier}, we propose Internal Latent Rotation, applying a randomized orthogonal matrix $R \in \mathbb{R}^{r \times r}$ to the latent factors. This exploits the rotational invariance of the factorization:
\begin{equation}
    W \approx \hat{U} \hat{V}^T = \hat{U} (R R^T) \hat{V}^T = (\hat{U} R) (\hat{V} R)^T = \tilde{U} \tilde{V}^T
\end{equation}
where $\tilde{U} = \hat{U} R$ and $\tilde{V} = \hat{V} R$ denote the rotated latent factors. Since $R$ is orthogonal ($R R^T = I$), the reconstruction remains exact in floating-point precision, but the coordinate distribution of the factors is fundamentally altered to be quantization-friendly.

\begin{theorem}[Delocalization via Rotation]
\label{thm:rotation}
Let $R$ be a random orthogonal matrix. By \textit{Levy's Lemma} (Concentration of Measure) \cite{ledoux2001concentration}, the mass of the rotated vectors $\tilde{U}$ concentrates around the expectation, forming a Gaussian-like distribution. This transformation maximizes the element-wise density, driving the expected distortion down to the \textbf{Gaussian Limit} (proof in Appendix \ref{app:proof_hierarchy}):
\begin{equation}
    \mathbb{E}[\lambda_{\text{Rot}}] \approx 1 - \frac{1}{r} \left( \sqrt{\frac{2r}{\pi}} \right)^2 = 1 - \frac{2}{\pi} \approx 0.36
\end{equation}
\end{theorem}

This rotation provides Coarse Alignment, stabilizing the Rank-1 scale approximation ($|U| \approx h \cdot l_u^T$). In standard SVD, high coherence forces shared scaling factors to accommodate outliers, overestimating the dynamic range for the majority of inliers. By enforcing an isotropic Gaussian distribution, we ensure that the magnitude matrix $|\tilde{U}|$ becomes nearly uniform. This homogenization allows the Rank-1 outer product to approximate the underlying geometry with reduced residual error, satisfying the structural constraints of the architecture.

Empirically, on the representative Llama-2 weight analyzed in Section \ref{subsec:internal_incoherence_processing}, this rotation reduces the mean distortion ($\lambda$) to 0.36, matching the theoretical limit, and suppresses the maximum distortion from 0.88 to 0.43. This confirms that scaling factors are no longer dominated by outliers.

\begin{figure}[t]
    \centering
    \includegraphics[width=1.0\linewidth]{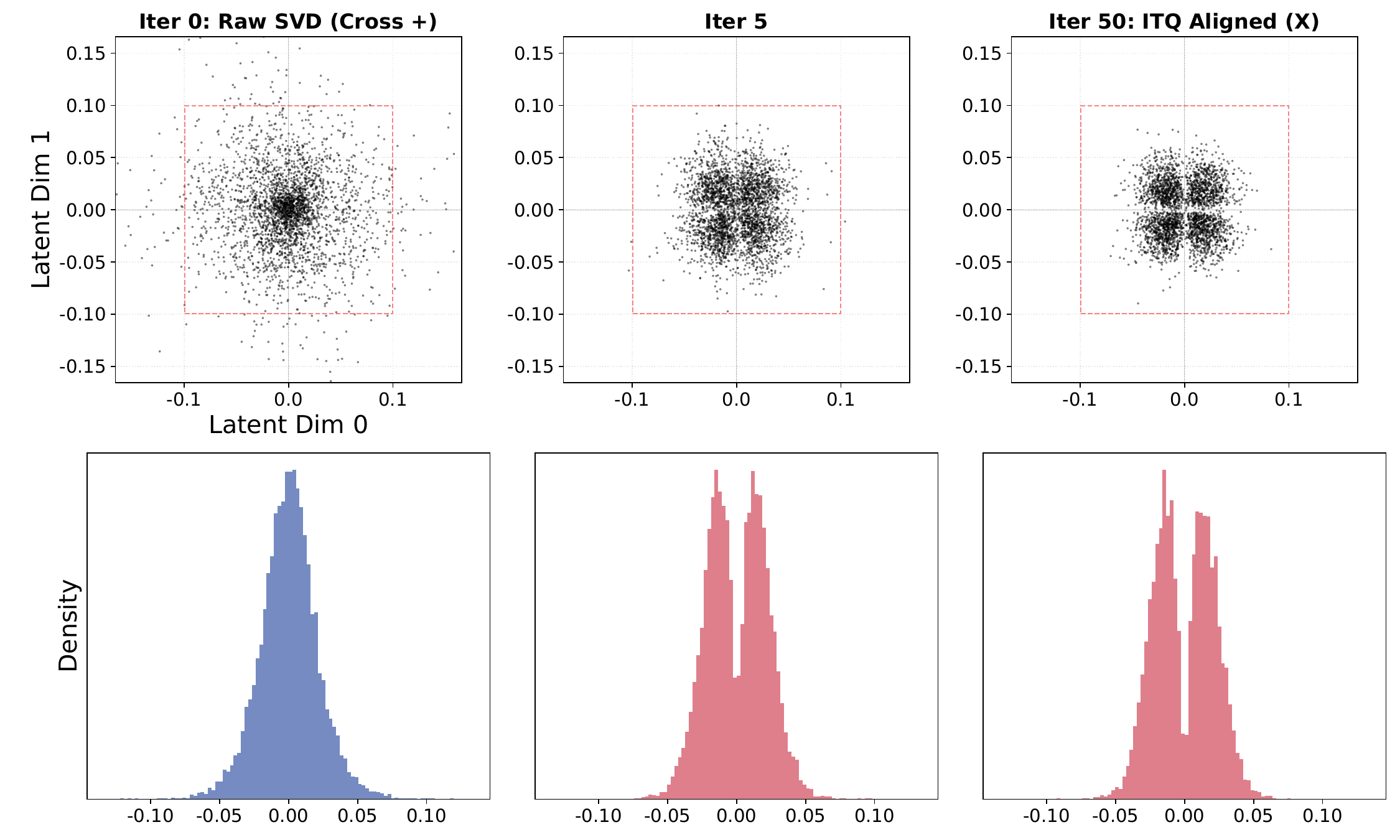}
    \caption{\textbf{Evolution of Latent Geometry via Joint-ITQ.} 
    Histograms of latent factors ($\hat{U}, \hat{V}$) from the Llama-2 7B 15th layer K projection (first two latent dimensions). 
    \textbf{(Left)} Raw SVD factors exhibit high coherence, concentrating probability mass near the zero decision boundary while containing significant outliers.
    \textbf{(Right)} Joint-ITQ (Iter 50) transforms this into a \textbf{bimodal distribution}, explicitly aligning the geometry with the binary vertices $\{\pm 1\}$ to maximize the decision margin.}
    \label{fig:itq_distribution_comparison}
    \vspace{-0.2in}
\end{figure}

\subsection{Joint-Iterative Latent Quantization}
\label{subsubsec:joint_itq}
While Random Rotation mitigates outliers by enforcing isotropy, it introduces rotational ambiguity. The resulting vectors often reside in the uncertainty zone near zero, misaligned with quantization targets. To minimize $\lambda$ ($\lambda \to 0$), we break this isotropy by formulating the alignment as a Joint Orthogonal Procrustes Problem \cite{gong2012iterative}. We seek a rotation $R^*$ minimizing the distance between the joint latent manifold $Z = [\hat{U}; \hat{V}]$ and binary vertices. We optimize a \textbf{shared} rotation $R$ over the concatenated manifold to align \textbf{both} latent factors simultaneously. As $R$ is orthogonal, this maintains reconstruction consistency $\hat{W} \approx (\hat{U}R)(\hat{V}R)^T = \hat{U}\hat{V}^T$ while minimizing quantization error for both factors.
\begin{equation}
    Z = \begin{bmatrix} \hat{U} \\ \hat{V} \end{bmatrix}
\end{equation}
Our objective is to find the rotation $R$ that best aligns this joint manifold with the vertices of the binary hypercube $B$. This is formalized as the following minimization problem: \begin{equation}
    \min_{R, B} \| B - Z R \|_F^2 \hspace{0.5em} \text{s.t.} \hspace{0.5em} R^T R = I_r, \hspace{0.5em} B \in \{\pm 1\}^{(d_{in} + d_{out}) \times r}
\end{equation}
We solve this non-convex objective via alternating minimization, iterating between the binary code update ($B = \text{sign}(ZR)$) and the rotation update via SVD on $B^T Z$ (Algorithm \ref{alg:joint_itq}).

This optimization transforms the unimodal Gaussian into a \textbf{Bimodal Distribution} aligned with the hypercube diagonals (Figure \ref{fig:itq_distribution_comparison}). By shifting mass from zero to $\{\pm 1\}$, it maximizes the Geometric Margin. On the representative Llama-2 weight analyzed in Section \ref{subsec:internal_incoherence_processing}, this further suppresses the mean distortion coefficient to \textbf{0.30}, surpassing the theoretical Gaussian limit ($\approx 0.36$). The process incurs negligible overhead, converging within 50 iterations with few seconds (Appendix \ref{subsec:itq_sensitivity}).

\begin{algorithm}[t]
   \small
   \caption{LittleBit-2 Initialization via Latent Geometric Alignment}
   \label{alg:joint_itq}
   \resizebox{0.95\linewidth}{!}{
     \begin{minipage}{\linewidth} 
\begin{algorithmic}[1]
   \STATE {\bfseries Input:} Weight $W$, Target Rank $r$, Iterations $T$
   \STATE {\bfseries Output:} Binary latent factors $\tilde{U}, \tilde{V}$, Scales $h, g, l$
   \STATE $U, \Sigma, V^T \leftarrow \text{SVD}(W)$ 
   \STATE $\hat{U} \leftarrow U_{:,1:r} \Sigma_{1:r}^{1/2}, \quad \hat{V} \leftarrow V_{:,1:r} \Sigma_{1:r}^{1/2}$ 
   \STATE $Z \leftarrow \text{Concat}([\hat{U} \,;\, \hat{V}])$ \hfill \textcolor{gray}{\textit{$\rhd$ $Z \in \mathbb{R}^{(d_{in} + d_{out}) \times r}$}} 
   \STATE Initialize $R \in \mathbb{R}^{r \times r}$ as random orthogonal matrix.
   \FOR{$t = 1$ {\bfseries to} $T$} 
       \STATE $B \leftarrow \text{sign}(Z R)$ \hfill \textcolor{gray}{\textit{$\rhd$ Project to Binary Vertices}}  
       \STATE $\Phi, \Omega, \Psi^T \leftarrow \text{SVD}(B^T Z)$ \hfill \textcolor{gray}{\textit{$\rhd$ Solve Procrustes Prob.}} 
       \STATE $R \leftarrow \Psi \Phi^T$ \hfill \textcolor{gray}{\textit{$\rhd$ Update Rotation}}
   \ENDFOR
   \STATE $\tilde{U} \leftarrow \hat{U} R, \quad \tilde{V} \leftarrow \hat{V} R$ \hfill \textcolor{gray}{\textit{$\rhd$ Rotate factors}}
   \STATE \textcolor{gray}{\textit{// Dual-SVID Init. (Scale extraction via Rank-1 SVD)}}
   \STATE $u_{\text{vec}}, u_{\text{rk}} \leftarrow \text{SVD}_1(|\tilde{U}|), \quad v_{\text{vec}}, v_{\text{rk}} \leftarrow \text{SVD}_1(|\tilde{V}|)$ 
     
   \STATE $h \leftarrow u_{\text{vec}}, \quad l \leftarrow u_{\text{rk}} \odot v_{\text{rk}}, \quad g \leftarrow v_{\text{vec}}$ \hfill \textcolor{gray}{\textit{$\rhd$ Set scales}}
   
   \STATE \textbf{return} $\tilde{U}, \tilde{V}, h, g, l$
\end{algorithmic}
\end{minipage}%
   }
\end{algorithm}

\section{Experiments}
\label{sec:experiments}
\begin{table*}[t]
\caption{\textbf{Main Results on Llama-2 and Llama-3.} Perplexity (PPL) on \textbf{WikiText-2} and average accuracy across \textbf{5 zero-shot tasks}. We denote \textbf{Body} (blocks only) and \textbf{Total} (including head and embedding) memory footprints in GB. At 0.1 bpp, LittleBit-2 compresses body weights to $<$1\%, leaving the footprint dominated by the fixed LM Head with Embedding. See Appendix \ref{app:extended_results} for detailed results.}
\label{tab:main_results_method_first}
\vspace{-0.1in}
\begin{center}
\begin{small}
\begin{sc}
\resizebox{0.97\textwidth}{!}{
\begin{tabular}{l|c|cccc|cccc|cccc}
\toprule
\multirow{3}{*}{\textbf{Method}} & \multirow{3}{*}{\textbf{Bits}} & \multicolumn{4}{c|}{\textbf{Llama-3 8B}} & \multicolumn{4}{c|}{\textbf{Llama-2 7B}} & \multicolumn{4}{c}{\textbf{Llama-2 13B}} \\
\cmidrule(lr){3-6} \cmidrule(lr){7-10} \cmidrule(lr){11-14}
 & & \multirow{2}{*}{\textbf{PPL}$\downarrow$} & \multirow{2}{*}{\textbf{Avg}$\uparrow$} & \multicolumn{2}{c|}{\textbf{Mem (GB)}} & \multirow{2}{*}{\textbf{PPL}$\downarrow$} & \multirow{2}{*}{\textbf{Avg$\uparrow$}} & \multicolumn{2}{c|}{\textbf{Mem (GB)}} & \multirow{2}{*}{\textbf{PPL}$\downarrow$} & \multirow{2}{*}{\textbf{Avg$\uparrow$}} & \multicolumn{2}{c}{\textbf{Mem (GB)}} \\
\cmidrule(lr){5-6} \cmidrule(lr){9-10} \cmidrule(lr){13-14}
 & & & & \scriptsize{Body {\tiny(\%)}} & \scriptsize{Total {\tiny(\%)}} & & & \scriptsize{Body {\tiny(\%)}} & \scriptsize{Total {\tiny(\%)}} & & & \scriptsize{Body {\tiny(\%)}} & \scriptsize{Total {\tiny(\%)}} \\
\midrule
\textit{FP16} & 16 & 6.10 & 72.97 & 14.0 \tiny{(100)} & 16.1 \tiny{(100)} & 5.47 & 68.88  & 13.0 \tiny{(100)} & 13.5 \tiny{(100)} & 4.88 & 71.87 & 25.4 \tiny{(100)} & 26.1 \tiny{(100)} \\
\midrule
GPTQ & 2 & 1480 & 36.01 & 2.0 \tiny{(14.2)} & 4.1 \tiny{(25.4)} & 52.2 & 38.79 & 1.8 \tiny{(14.2)} & 2.4 \tiny{(17.5)} & 23.6 & 44.34 & 3.6 \tiny{(14.1)} & 4.2 \tiny{(16.3)} \\

EfficientQAT & 2 & 9.80 & 63.49 & 2.0 \tiny{(14.2)} & 4.1 \tiny{(25.4)} & 7.17 & 62.93 & 1.8 \tiny{(14.2)} & 2.4 \tiny{(17.5)} & 6.08 & 68.06 & 3.6 \tiny{(14.1)} & 4.2 \tiny{(16.3)} \\
\midrule
\multicolumn{14}{c}{\textit{1-bit Regime}} \\
\midrule
BiLLM & 1.1 & 59.37 & 37.73 & 2.5 \tiny{(18.2)} & 4.6 \tiny{(28.9)} & 29.00 & 40.81 & 2.4 \tiny{(18.2)} & 2.9 \tiny{(21.4)} & 21.53 & 49.12 & 4.6 \tiny{(18.1)} & 5.3 \tiny{(20.2)} \\
ARB-LLM & 1.1 & 27.63 & 52.75 & 2.4 \tiny{(17.5)} & 4.5 \tiny{(28.2)} & 15.88 & 52.32 & 2.3 \tiny{(17.5)} & 2.8 \tiny{(20.7)} & 12.13 & 57.98 & 4.5 \tiny{(17.8)} & 5.1 \tiny{(19.5)} \\
OneBit & 1.0 & 13.09 & 52.23 & 0.9 \tiny{(6.4)} & 3.0 \tiny{(18.6)} & 8.36 & 54.72 & 0.8 \tiny{(6.4)} & 1.4 \tiny{(10.0)} & 7.41 & 58.47 & 1.6 \tiny{(6.4)} & 2.3 \tiny{(8.7)} \\
LittleBit & 1.0 & 16.30 & 47.11 & 0.9 \tiny{(6.3)} & 3.0 \tiny{(18.6)} & 9.08 & 52.81 & 0.8 \tiny{(6.3)} & 1.3 \tiny{(10.0)} & 8.18 & 51.49 & 1.6 \tiny{(6.3)} & 2.3 \tiny{(8.7)} \\
\textbf{LittleBit-2 (Ours)} & 1.0 & \textbf{11.53} & \textbf{57.33} & 0.9 \tiny{(6.3)} & 3.0 \tiny{(18.6)} & \textbf{8.27} & \textbf{55.79} & 0.8 \tiny{(6.3)} & 1.3 \tiny{(10.0)} & \textbf{7.37} & \textbf{60.53} & 1.6 \tiny{(6.3)} & 2.3 \tiny{(8.7)} \\
\midrule
\multicolumn{14}{c}{\textit{Sub-1-bit Regime (Extreme Compression)}} \\
\midrule

STBLLM & 0.55 & 241.96 & 36.76 & - & - & 32.93 & 43.40 & - & - & 27.05 & 45.62 & - & - \\
LittleBit & 0.55 & 18.47 & 44.59 & 0.5 \tiny{(3.5)} & 2.6 \tiny{(16.1)} & 10.47 & 48.49 & 0.5 \tiny{(3.6)} & 1.0 \tiny{(7.3)} & 9.24 & 51.76 & 0.9 \tiny{(3.5)} & 1.6 \tiny{(5.9)} \\
\textbf{LittleBit-2 (Ours)} & 0.55 & \textbf{14.01} & \textbf{51.33} & 0.5 \tiny{(3.5)} & 2.6 \tiny{(16.1)} & \textbf{9.65} & \textbf{50.97} & 0.5 \tiny{(3.6)} & 1.0 \tiny{(7.3)} & \textbf{8.56} & \textbf{55.64} & 0.9 \tiny{(3.5)} & 1.6 \tiny{(5.9)} \\
\cmidrule{1-14}
LittleBit & 0.1 & 26.11 & \textbf{41.80} & 0.1 \tiny{(0.7)} & 2.2 \tiny{(13.7)} & 15.92 & 41.33 & 0.1 \tiny{(0.7)} & 0.6 \tiny{(4.6)} & 15.09 & 40.99 & 0.2 \tiny{(0.7)} & 0.8 \tiny{(3.2)} \\
\textbf{LittleBit-2 (Ours)} & 0.1 & \textbf{23.74} & 41.64 & 0.1 \tiny{(0.7)} & 2.2 \tiny{(13.7)} & \textbf{14.70} & \textbf{42.06} & 0.1 \tiny{(0.7)} & 0.6 \tiny{(4.6)} & \textbf{13.94} & \textbf{42.18} & 0.2 \tiny{(0.7)} & 0.8 \tiny{(3.2)} \\
\bottomrule
\end{tabular}
}
\end{sc}
\end{small}
\end{center}
\vspace{-0.2in}
\end{table*}

\subsection{Synthetic Validation}
\label{subsec:synthetic_validation}
\begin{figure}[t]
\begin{center}
\includegraphics[width=\linewidth]{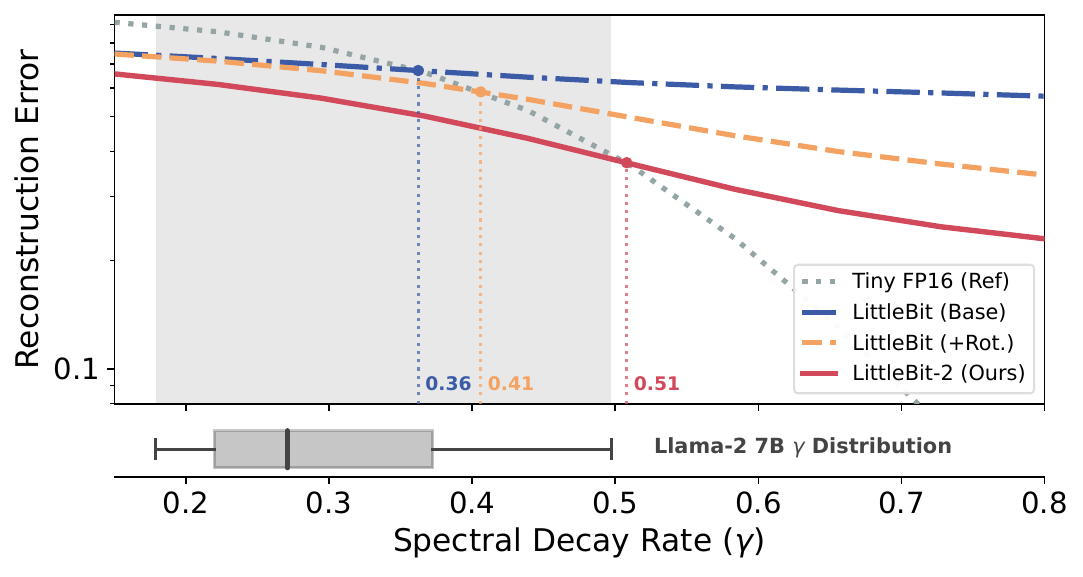}
\caption{
\textbf{Spectral Break-Even Analysis.} 
\textbf{(Top)} Reconstruction error (MSE) vs. spectral decay rate $\gamma$. LittleBit-2 (with Joint-ITQ) extends the range where 1-bit quantization outperforms FP16.
\textbf{(Bottom)} Distribution of $\gamma$ in Llama-2 7B weights. The shaded region (5th--95th percentile) shows real-world weights fall mostly within the heavy-tailed regime. LittleBit-2 dominates this region, whereas the baseline fails in the upper quantile.}
\label{fig:synthetic}
\end{center}
\vspace{-0.2in}
\end{figure}

To validate Proposition \ref{thm:superiority}, we conducted reconstruction experiments on synthetic weights. We generated random matrices $W \in \mathbb{R}^{4096 \times 4096}$ with singular values decaying according to a power-law $\sigma_k \propto k^{-\gamma}$, varying the decay rate $\gamma \in [0.1, 0.8]$. We compared the reconstruction fidelity of \textbf{LittleBit-2} against baselines under identical memory constraints. The approximation employs initialized 1-bit quantization to measure the reconstruction error, whereas Tiny-Rank FP16 is obtained via truncated SVD.

Figure \ref{fig:synthetic} (Top) illustrates the reconstruction error (MSE) as a function of $\gamma$. We observe a distinct phase transition governed by the spectral shape. While standard LittleBit outperforms Tiny-Rank FP16 only in the heavy-tailed regime ($\gamma \lesssim 0.36$), our geometric interventions progressively extend this superiority. Specifically, applying internal rotation pushes the break-even point to $\gamma \approx 0.41$, and \textbf{LittleBit-2} (with Joint-ITQ) further extends it to $\gamma \approx 0.51$. This confirms that for heavy-tailed spectra, the information gain from massive rank expansion ($r_{\text{bin}} \approx 16 r_{\text{fp}}$) outweighs the precision loss from 1-bit quantization.

To confirm that this expanded effective range covers the operational regime of modern LLMs, Figure \ref{fig:synthetic} (Bottom) compares the empirical decay rates from \textbf{all linear layers of Llama-2 7B} against our method's performance. Here, all gammas are calculated by log linear regression of real weights. The majority (90\%) of observed rates lie within $\gamma \in [0.19, 0.47]$ (median: $0.27$), a range that completely overlaps with the 1-bit LittleBit-2's optimal reconstruction zone. This confirms that the heavy-tailed spectral properties of LLMs are intrinsically suited for sub-1-bit compression.

\subsection{Main Results}
\label{subsec:main_results}
\paragraph{Experimental Setup.}
We evaluate our method on Llama-2 7B, 13B \cite{touvron2023llama} and Llama-3 8B \cite{grattafiori2024llama}. For evaluation metrics, we measure the perplexity (PPL) on WikiText-2 \cite{merity2016pointer} using a sequence length of 2048. To assess reasoning capabilities, we report the average zero-shot accuracy across five benchmark tasks: HellaSwag \cite{zellers2019hellaswag}, ARC-Easy, ARC-Challenge \cite{clark2018think}, PIQA \cite{bisk2020piqa}, and Winogrande \cite{sakaguchi2021winogrande}.

\paragraph{Baselines.}
We compare LittleBit-2 against a wide range of quantization methods. For \textbf{GPTQ} \cite{frantar2022gptq}, \textbf{EfficientQAT} \cite{chen2025efficientqat}, \textbf{STBLLM} \cite{dong2024stbllm}, \textbf{BiLLM} \cite{huang2024billm}, and \textbf{ARB-LLM} \cite{li2024arb} we utilized their officially released codes to reproduce the results. Note that all these baselines utilize a group size of 128. To ensure a rigorous comparison with \textbf{LittleBit} \cite{lee2025littlebit} and \textbf{OneBit} \cite{xu2024onebit}, we adopted the identical training configuration (including the number of tokens, epochs, and learning rate) as detailed in \cite{lee2025littlebit}.

\paragraph{Performance Analysis.}
Table \ref{tab:main_results_method_first} summarizes the performance comparison. LittleBit-2 establishes a new state-of-the-art in the low-bit quantization regime. In the 1-bit setting, LittleBit-2 significantly outperforms binary quantization methods such as BiLLM and ARB-LLM. LittleBit-2 achieves a perplexity of \textbf{11.53} on Llama-3 8B, significantly surpassing the baseline LittleBit (16.30). Our method matches or surpasses the performance of OneBit (11.53 vs. 13.09 in PPL, 57.33\% vs. 52.33\% in Avg).

We observe a scaling anomaly in the baseline LittleBit, where Llama-2 13B underperforms Llama-2 7B on zero-shot tasks (51.49\% vs. 52.81\%). This suggests that standard initialization struggles to handle the increased optimization difficulty of larger models. LittleBit-2 successfully resolves this issue, restoring the expected performance scaling (60.53\% vs. 55.79\%). This confirms that latent geometry alignment provides the necessary stability for training larger-scale models.

The superiority of LittleBit-2 becomes most pronounced in the deep compression regime. As the bit-width decreases to 0.55 bpp, STBLLM exhibits severe performance collapse (PPL 241.96 on Llama-3). While the baseline LittleBit maintains functionality, it still suffers from optimization instability (PPL 18.47). In contrast, LittleBit-2 retains high fidelity (PPL 14.01), reducing the perplexity gap with the FP16 model significantly. 

\textbf{Even in the extreme 0.1 bpp regime, compressing the model body to $<$1\% of its original size ($\sim$0.1 GB for 8B models), LittleBit-2 remains functional}. It achieves a perplexity of 23.74 on Llama-3 8B, outperforming the baseline LittleBit (PPL 26.11). This result aligns with the theoretical prediction in Section \ref{sec:method}. LittleBit-2 effectively retains the heavy-tailed information essential for minimal functionality.

\paragraph{Scalability on Large-Scale Models}
\label{subsec:gemma_exp}
To verify scalability on larger, recent architectures, we extended evaluation to \textbf{Gemma-3 27B}. We focused on the extreme compression regime (0.1 bpp) to test the limits of latent information retention.

As summarized in Table \ref{tab:gemma_results}, \textbf{LittleBit-2} demonstrates robust scalability even in the extreme compression regime. Notably, in the 0.1 bpp setting, while the Tiny-Rank FP approximation collapses (PPL $>$ 35), LittleBit-2 maintains functional capabilities with a PPL of 16.38 and an average zero-shot accuracy of 47.06\%, significantly outperforming the baseline. 

\begin{table}[h]
\caption{\textbf{Results on Gemma-3 27B.} Comparison of Perplexity (PPL) and average zero-shot accuracy (Avg). The evaluation focuses on the extreme 0.1-bit regimes. LittleBit-2 consistently outperforms the baselines, showing gains in the 0.1 bpp setting.}
\label{tab:gemma_results}
\vspace{-0.1in}
\begin{center}
\begin{small}
\begin{sc}
\resizebox{0.60\columnwidth}{!}{
\begin{tabular}{l|c|cc}
\toprule
\textbf{Method} & \textbf{Bits} & \textbf{PPL}$\downarrow$ & \textbf{Avg (\%)}$\uparrow$ \\
\midrule
\textit{FP16 (Ref)} & 16.0 & 4.58 & 79.48 \\
\midrule
FP (Tiny-Rank) & 0.1 & 37.53 & 39.15 \\
LittleBit & 0.1 & 17.72 & 45.52 \\
\textbf{LittleBit-2 (Ours)} & \textbf{0.1} & \textbf{16.38} & \textbf{47.06} \\
\bottomrule
\end{tabular}
\vspace{-0.2in}
}
\end{sc}
\end{small}
\end{center}
\end{table}

\subsection{Ablation Study}
\label{subsec:ablation}
We conduct a component-wise ablation study on Llama-3 8B across standard (1.0 bpp) and extreme (0.1 bpp) regimes to isolate geometric alignment contributions. Table \ref{tab:ablation} dissects the performance gains. First, the collapse of Tiny-Rank FP (PPL 59.44) at 0.1 bpp contrasts with the stability of LittleBit Base (PPL 26.11). This empirically validates the \textbf{Spectral Break-Even Condition} (Proposition \ref{thm:superiority}), confirming that in the heavy-tailed spectral regime, the information gain from massive rank expansion significantly outweighs the precision loss incurred by binarization. Second, geometric alignment yields progressive improvements. While Random Rotation reduces error by mitigating outliers ($16.30 \to 12.63$), \textbf{LittleBit-2} (Joint-ITQ) achieves the best performance (11.53) by explicitly aligning latent factors with the binary hypercube vertices, thereby maximizing the geometric decision margin.
\begin{table}[h]
\caption{\textbf{Ablation Study on Llama-3 8B.} Comparison of WikiText-2 Perplexity (PPL). The results highlight the failure of Tiny-Rank FP16 in extreme regimes (0.1 bpp) and the progressive improvement from geometric alignment (Rotation $\to$ ITQ).}
\label{tab:ablation}
\vspace{-0.1in}
\begin{center}
\begin{small}
\begin{sc}
\resizebox{0.60\columnwidth}{!}{
\begin{tabular}{l|cc}
\toprule
\textbf{Method} & \textbf{0.1 bpp} & \textbf{1.0 bpp} \\
\midrule
\textit{Original Model (FP16)} & \multicolumn{2}{c}{\textit{6.10}} \\
\midrule
FP (Tiny-Rank) & 59.44 & 26.24 \\
LittleBit (Base) & 26.11 & 16.30 \\
+ Random Rotation & 24.16 & 12.63 \\
\textbf{LittleBit-2 (Ours)} & \textbf{23.74} & \textbf{11.53} \\
\bottomrule
\end{tabular}
}
\end{sc}
\end{small}
\end{center}
\vspace{-0.2in}
\end{table}

\begin{figure}[t]
    \begin{center}
        \includegraphics[width=\linewidth]{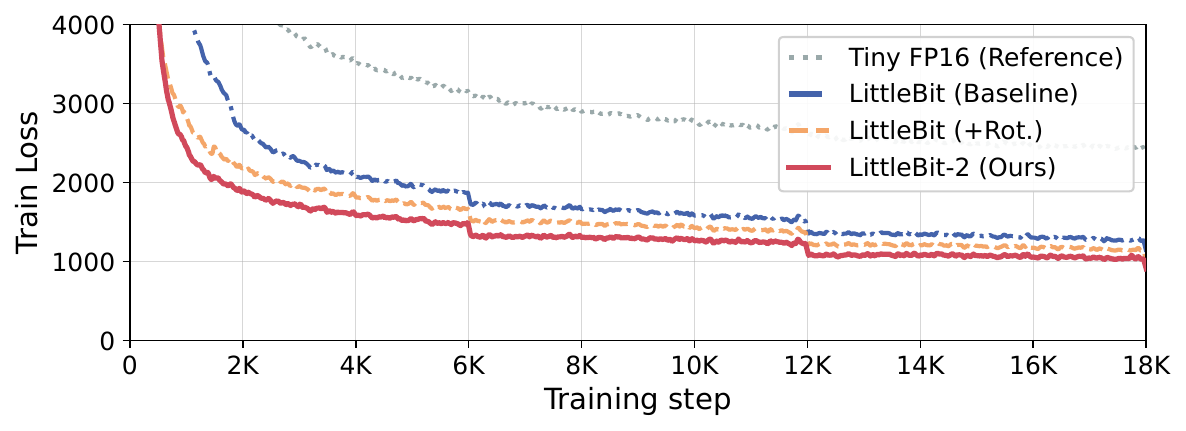}
        \vspace{-0.1in}
        \caption{\textbf{Training Convergence.} Comparison of training loss trajectories for 1-bit QAT on Llama-2 7B.
        \textbf{Tiny-Rank FP16} (Gray) suffers from high error.
        While standard \textbf{LittleBit} (Blue) captures more capacity, it exhibits slower convergence due to latent misalignment.
        Applying \textbf{Internal Rotation} (Orange) improves optimization speed by mitigating outliers, and \textbf{LittleBit-2} (Red, with Joint-ITQ) achieves the fastest convergence and lowest final loss.
        }
        \label{fig:dynamics}
    \end{center}
    \vspace{-0.2in}
\end{figure}

\section{Discussion}
\label{sec:discussion}
\subsection{Training Stability}
\label{subsec:stability_mechanism}
We analyze the training dynamics to understand why geometric alignment leads to superior performance. Figure \ref{fig:dynamics} illustrates the training loss trajectories of 1-bit quantization on Llama-2 7B. The \textbf{Tiny-Rank FP16} baseline saturates early at a high loss plateau, confirming the fundamental bottleneck of minimal rank. In contrast, \textbf{LittleBit-2} achieves the lowest final loss and accelerates convergence in the early training phase compared to the baseline.

\begin{figure}[t]
    \centering
    \includegraphics[width=1.0\linewidth]{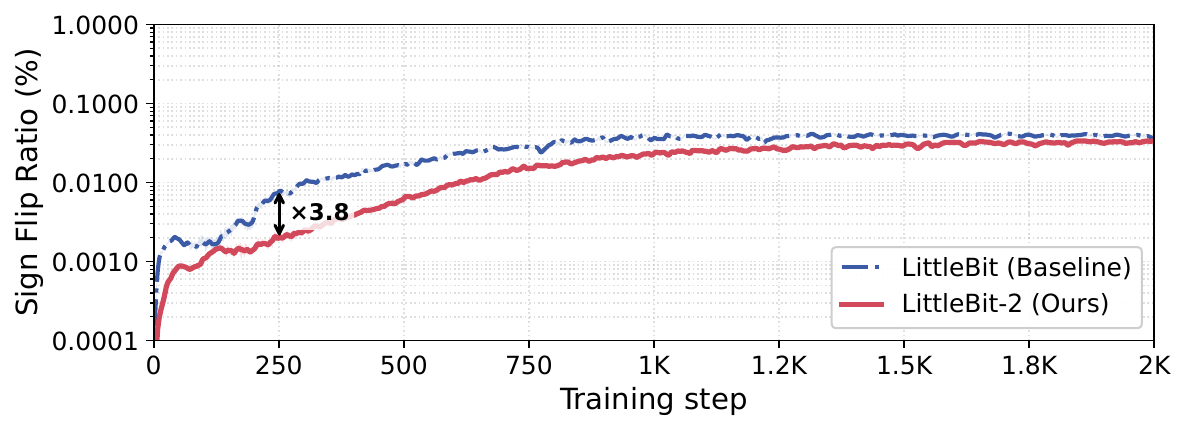}
    \vspace{-0.1in}
    \caption{\textbf{Sign Flipping Analysis.} The percentage of binary parameters changing signs per step during the first 2k steps (7\% of total) under the same setting as Figure \ref{fig:dynamics}. LittleBit-2 (Blue) significantly reduces oscillation compared to the baseline (Red). This confirms that in the early stage of training geometric margin stabilizes the weights against stochastic gradient noise.}
    \label{fig:flipping_ratio}
    \vspace{-0.2in}
\end{figure}

To investigate this acceleration, we analyze the Sign Flipping Ratio, defined as the percentage of binary parameters changing state ($\pm 1$) per step. As shown in Figure \ref{fig:flipping_ratio}, the standard LittleBit baseline exhibits a persistently high flipping rate. This indicates that early in training, many latent parameters are trapped near the decision boundary ($x=0$), oscillating due to gradient noise rather than learning features.

We attribute the stability of LittleBit-2 to the Geometric Margin induced by Joint-ITQ. While standard random rotation yields a unimodal Gaussian distribution concentrated at zero (maximizing instability), Joint-ITQ enforces a bimodal separation aligned with the binary vertices. By pushing latent factors away from the unstable zero-crossing region, our method effectively locks the weights against high-frequency gradient noise. This geometric preconditioning stabilizes the optimization landscape, enabling effective feature learning from the initial iterations.

\subsection{Inference Efficiency}
\label{subsec:complexity}
While Latent Geometry Alignment improves quantization accuracy, it does not compromise inference efficiency. LittleBit-2 retains the \textit{identical} inference architecture as the original LittleBit framework \cite{lee2025littlebit}. Consequently, our method directly inherits the established computational advantages of the Low-Rank Binary structure.

The core advantage of this architecture lies in its MatMul-free design. By decomposing weights into low-rank binary factors, it replaces FP16 GEMV operations with Bitwise Operations (BOPs), while reducing the total parameter count. For instance, a Llama-2 7B MLP layer at 0.3 bpp reduces computational complexity from 90.2M FLOPs (add+mul) to 13M FLOPs (add) with 13M sign flips.

Empirical benchmarks reported in prior work \cite{lee2025littlebit} further validate this efficiency on hardware. At the kernel level, a Llama-2 70B MLP layer compressed to 0.1 bpp demonstrates an \textbf{11.6$\times$ speedup} (0.288 ms $\to$ 0.025 ms) compared to the cuBLAS FP16 baseline. This efficiency translates to substantial gains in end-to-end generation; a 0.1 bpp Llama-2 7B model achieves a \textbf{2.46$\times$ speedup} (82.6 tokens/s $\to$ 203.2 tokens/s) in a 128-token decoding, batch 1 setting. Since LittleBit-2 shares this underlying computational graph, it guarantees the same level of acceleration.

\section{Conclusion}
We present \textbf{LittleBit-2}, a framework that realizes the spectral gain in sub-1-bit LLMs by resolving Latent Geometry Misalignment via Joint-ITQ. By transforming coherent singular vectors into quantization-friendly bimodal distributions, our approach maximizes the geometric decision margin and realizes the information-theoretic advantage of the Low-Rank Binary strategy. LittleBit-2 achieves state-of-the-art performance down to 0.1 bpp, effectively matching 1-bit baselines and confirming that geometrically aligned extreme compression is a viable path for deploying foundation models on memory-constrained edge devices.

Future directions include developing adaptive rank allocation guided by spectral decay ($\gamma$), exploring hybrid architectures combining FP components with LittleBit, and employing advanced optimization strategies to further accelerate training.

\section*{Impact Statement}
We present a method for extreme model compression that significantly lowers the computational and energy costs of deploying LLMs. This work has positive societal impacts by promoting energy-efficient computing and democratizing access to state-of-the-art AI technology. While deploying powerful models on edge devices introduces challenges regarding the control of potential misuse, it also offers significant benefits for data privacy.

\bibliography{ref}

@article{grattafiori2024llama,
  title={The llama 3 herd of models},
  author={Grattafiori, Aaron and Dubey, Abhimanyu and Jauhri, Abhinav and Pandey, Abhinav and Kadian, Abhishek and Al-Dahle, Ahmad and Letman, Aiesha and Mathur, Akhil and Schelten, Alan and Vaughan, Alex and others},
  journal={arXiv preprint arXiv:2407.21783},
  year={2024}
}

@article{martin2021implicit,
  title={Implicit self-regularization in deep neural networks: Evidence from random matrix theory and implications for learning},
  author={Martin, Charles H and Mahoney, Michael W},
  journal={Journal of Machine Learning Research},
  volume={22},
  number={165},
  pages={1--73},
  year={2021}
}

@article{brown2020language,
  title={Language models are few-shot learners},
  author={Brown, Tom and Mann, Benjamin and Ryder, Nick and Subbiah, Melanie and Kaplan, Jared D and Dhariwal, Prafulla and Neelakantan, Arvind and Shyam, Pranav and Sastry, Girish and Askell, Amanda and others},
  journal={Advances in neural information processing systems},
  volume={33},
  pages={1877--1901},
  year={2020}
}

@article{touvron2023llama,
  title={Llama: Open and efficient foundation language models},
  author={Touvron, Hugo and Lavril, Thibaut and Izacard, Gautier and Martinet, Xavier and Lachaux, Marie-Anne and Lacroix, Timoth{\'e}e and Rozi{\`e}re, Baptiste and Goyal, Naman and Hambro, Eric and Azhar, Faisal and others},
  journal={arXiv preprint arXiv:2302.13971},
  year={2023}
}

@article{merity2016pointer,
  title={Pointer sentinel mixture models},
  author={Merity, Stephen and Xiong, Caiming and Bradbury, James and Socher, Richard},
  journal={arXiv preprint arXiv:1609.07843},
  year={2016}
}

@incollection{gholami2022survey,
  title={A survey of quantization methods for efficient neural network inference},
  author={Gholami, Amir and Kim, Sehoon and Dong, Zhen and Yao, Zhewei and Mahoney, Michael W and Keutzer, Kurt},
  booktitle={Low-power computer vision},
  pages={291--326},
  year={2022},
  publisher={Chapman and Hall/CRC}
}

@article{frantar2022gptq,
  title={Gptq: Accurate post-training quantization for generative pre-trained transformers},
  author={Frantar, Elias and Ashkboos, Saleh and Hoefler, Torsten and Alistarh, Dan},
  journal={arXiv preprint arXiv:2210.17323},
  year={2022}
}

@article{lin2024awq,
  title={Awq: Activation-aware weight quantization for on-device llm compression and acceleration},
  author={Lin, Ji and Tang, Jiaming and Tang, Haotian and Yang, Shang and Chen, Wei-Ming and Wang, Wei-Chen and Xiao, Guangxuan and Dang, Xingyu and Gan, Chuang and Han, Song},
  journal={Proceedings of machine learning and systems},
  volume={6},
  pages={87--100},
  year={2024}
}

@article{wang2023bitnet,
  title={Bitnet: Scaling 1-bit transformers for large language models},
  author={Wang, Hongyu and Ma, Shuming and Dong, Li and Huang, Shaohan and Wang, Huaijie and Ma, Lingxiao and Yang, Fan and Wang, Ruiping and Wu, Yi and Wei, Furu},
  journal={arXiv preprint arXiv:2310.11453},
  year={2023}
}

@article{xu2024onebit,
  title={Onebit: Towards extremely low-bit large language models},
  author={Xu, Yuzhuang and Han, Xu and Yang, Zonghan and Wang, Shuo and Zhu, Qingfu and Liu, Zhiyuan and Liu, Weidong and Che, Wanxiang},
  journal={Advances in Neural Information Processing Systems},
  volume={37},
  pages={66357--66382},
  year={2024}
}

@article{lee2025littlebit,
  title={LittleBit: Ultra Low-Bit Quantization via Latent Factorization},
  author={Lee, Banseok and Kim, Dongkyu and You, Youngcheon and Kim, Youngmin},
  journal={arXiv preprint arXiv:2506.13771},
  year={2025}
}

@article{huang2024billm,
  title={Billm: Pushing the limit of post-training quantization for llms},
  author={Huang, Wei and Liu, Yangdong and Qin, Haotong and Li, Ying and Zhang, Shiming and Liu, Xianglong and Magno, Michele and Qi, Xiaojuan},
  journal={arXiv preprint arXiv:2402.04291},
  year={2024}
}

@article{li2024arb,
  title={Arb-llm: Alternating refined binarizations for large language models},
  author={Li, Zhiteng and Yan, Xianglong and Zhang, Tianao and Qin, Haotong and Xie, Dong and Tian, Jiang and Kong, Linghe and Zhang, Yulun and Yang, Xiaokang and others},
  journal={arXiv preprint arXiv:2410.03129},
  year={2024}
}

@article{dong2024stbllm,
  title={Stbllm: Breaking the 1-bit barrier with structured binary llms},
  author={Dong, Peijie and Li, Lujun and Zhong, Yuedong and Du, Dayou and Fan, Ruibo and Chen, Yuhan and Tang, Zhenheng and Wang, Qiang and Xue, Wei and Guo, Yike and others},
  journal={arXiv preprint arXiv:2408.01803},
  year={2024}
}

@article{ma2024era,
  title={The era of 1-bit llms: All large language models are in 1.58 bits},
  author={Ma, Shuming and Wang, Hongyu and Ma, Lingxiao and Wang, Lei and Wang, Wenhui and Huang, Shaohan and Dong, Li and Wang, Ruiping and Xue, Jilong and Wei, Furu},
  journal={arXiv preprint arXiv:2402.17764},
  year={2024}
}

@inproceedings{liu2024llm,
  title={Llm-qat: Data-free quantization aware training for large language models},
  author={Liu, Zechun and Oguz, Barlas and Zhao, Changsheng and Chang, Ernie and Stock, Pierre and Mehdad, Yashar and Shi, Yangyang and Krishnamoorthi, Raghuraman and Chandra, Vikas},
  booktitle={Findings of the Association for Computational Linguistics: ACL 2024},
  pages={467--484},
  year={2024}
}

@article{du2024bitdistiller,
  title={Bitdistiller: Unleashing the potential of sub-4-bit llms via self-distillation},
  author={Du, Dayou and Zhang, Yijia and Cao, Shijie and Guo, Jiaqi and Cao, Ting and Chu, Xiaowen and Xu, Ningyi},
  journal={arXiv preprint arXiv:2402.10631},
  year={2024}
}

@article{bengio2013estimating,
  title={Estimating or propagating gradients through stochastic neurons for conditional computation},
  author={Bengio, Yoshua and L{\'e}onard, Nicholas and Courville, Aaron},
  journal={arXiv preprint arXiv:1308.3432},
  year={2013}
}

@inproceedings{chen2025efficientqat,
  title={Efficientqat: Efficient quantization-aware training for large language models},
  author={Chen, Mengzhao and Shao, Wenqi and Xu, Peng and Wang, Jiahao and Gao, Peng and Zhang, Kaipeng and Luo, Ping},
  booktitle={Proceedings of the 63rd Annual Meeting of the Association for Computational Linguistics (Volume 1: Long Papers)},
  pages={10081--10100},
  year={2025}
}

@article{tseng2024quip,
  title={Quip\#: Even better llm quantization with hadamard incoherence and lattice codebooks},
  author={Tseng, Albert and Chee, Jerry and Sun, Qingyao and Kuleshov, Volodymyr and De Sa, Christopher},
  journal={arXiv preprint arXiv:2402.04396},
  year={2024}
}

@article{ashkboos2024quarot,
  title={Quarot: Outlier-free 4-bit inference in rotated llms},
  author={Ashkboos, Saleh and Mohtashami, Amirkeivan and Croci, Maximilian L and Li, Bo and Cameron, Pashmina and Jaggi, Martin and Alistarh, Dan and Hoefler, Torsten and Hensman, James},
  journal={Advances in Neural Information Processing Systems},
  volume={37},
  pages={100213--100240},
  year={2024}
}

@article{liu2024spinquant,
  title={Spinquant: Llm quantization with learned rotations},
  author={Liu, Zechun and Zhao, Changsheng and Fedorov, Igor and Soran, Bilge and Choudhary, Dhruv and Krishnamoorthi, Raghuraman and Chandra, Vikas and Tian, Yuandong and Blankevoort, Tijmen},
  journal={arXiv preprint arXiv:2405.16406},
  year={2024}
}

@article{zellers2019hellaswag,
  title={Hellaswag: Can a machine really finish your sentence?},
  author={Zellers, Rowan and Holtzman, Ari and Bisk, Yonatan and Farhadi, Ali and Choi, Yejin},
  journal={arXiv preprint arXiv:1905.07830},
  year={2019}
}

@article{clark2018think,
  title={Think you have solved question answering? try arc, the ai2 reasoning challenge},
  author={Clark, Peter and Cowhey, Isaac and Etzioni, Oren and Khot, Tushar and Sabharwal, Ashish and Schoenick, Carissa and Tafjord, Oyvind},
  journal={arXiv preprint arXiv:1803.05457},
  year={2018}
}

@inproceedings{bisk2020piqa,
  title={Piqa: Reasoning about physical commonsense in natural language},
  author={Bisk, Yonatan and Zellers, Rowan and Gao, Jianfeng and Choi, Yejin and others},
  booktitle={Proceedings of the AAAI conference on artificial intelligence},
  volume={34},
  number={05},
  pages={7432--7439},
  year={2020}
}

@article{sakaguchi2021winogrande,
  title={Winogrande: An adversarial winograd schema challenge at scale},
  author={Sakaguchi, Keisuke and Bras, Ronan Le and Bhagavatula, Chandra and Choi, Yejin},
  journal={Communications of the ACM},
  volume={64},
  number={9},
  pages={99--106},
  year={2021},
  publisher={ACM New York, NY, USA}
}

@article{gong2012iterative,
  title={Iterative quantization: A procrustean approach to learning binary codes for large-scale image retrieval},
  author={Gong, Yunchao and Lazebnik, Svetlana and Gordo, Albert and Perronnin, Florent},
  journal={IEEE transactions on pattern analysis and machine intelligence},
  volume={35},
  number={12},
  pages={2916--2929},
  year={2012},
  publisher={IEEE}
}

@book{ledoux2001concentration,
  title={The concentration of measure phenomenon},
  author={Ledoux, Michel},
  number={89},
  year={2001},
  publisher={American Mathematical Soc.}
}

@article{eckart1936approximation,
  title={The approximation of one matrix by another of lower rank},
  author={Eckart, Carl and Young, Gale},
  journal={Psychometrika},
  volume={1},
  number={3},
  pages={211--218},
  year={1936},
  publisher={Springer-Verlag}
}

@article{kim2019qkd,
  title={Qkd: Quantization-aware knowledge distillation},
  author={Kim, Jangho and Bhalgat, Yash and Lee, Jinwon and Patel, Chirag and Kwak, Nojun},
  journal={arXiv preprint arXiv:1911.12491},
  year={2019}
}

@article{han2015deep,
  title={Deep compression: Compressing deep neural networks with pruning, trained quantization and huffman coding},
  author={Han, Song and Mao, Huizi and Dally, William J},
  journal={arXiv preprint arXiv:1510.00149},
  year={2015}
}
\bibliographystyle{icml2026}

\newpage
\appendix
\onecolumn
\section{Detailed Theoretical Analysis}
\label{app:proofs}
\subsection{Proof of Lemma \ref{lemma:distortion_duality} (Distortion-Geometry Duality)}
\label{app:proof_distortion}
Let $\mathbf{b} = \text{sign}(\mathbf{u}) \in \{-1, +1\}^r$. The objective is to minimize $f(\alpha) = \| \mathbf{u} - \alpha \mathbf{b} \|_2^2$.
Differentiating with respect to $\alpha$:
\begin{equation}
    \frac{\partial f}{\partial \alpha} = -2 \mathbf{b}^T (\mathbf{u} - \alpha \mathbf{b}) = -2 (\mathbf{b}^T \mathbf{u} - \alpha \mathbf{b}^T \mathbf{b}) = 0
\end{equation}
Solving for $\alpha$, and noting that $\mathbf{b}^T \mathbf{b} = r$ and $\mathbf{b}^T \mathbf{u} = \sum |u_i| = \| \mathbf{u} \|_1$:
\begin{equation}
    \alpha^* = \frac{\mathbf{b}^T \mathbf{u}}{\mathbf{b}^T \mathbf{b}} = \frac{\| \mathbf{u} \|_1}{r}
\end{equation}
Substituting $\alpha^*$ back into the normalized error function $\lambda(\mathbf{u}) = \mathcal{E}(\mathbf{u})/\|\mathbf{u}\|_2^2$:
\begin{equation}
    \mathcal{E}(\mathbf{u}) = \| \mathbf{u} \|_2^2 - 2\alpha^* (\mathbf{b}^T \mathbf{u}) + (\alpha^*)^2 (\mathbf{b}^T \mathbf{b}) = \| \mathbf{u} \|_2^2 - \frac{\| \mathbf{u} \|_1^2}{r}
\end{equation}
\begin{equation}
    \therefore \lambda(\mathbf{u}) = 1 - \frac{1}{r} \left( \frac{\| \mathbf{u} \|_1}{\| \mathbf{u} \|_2} \right)^2 \quad \blacksquare
\end{equation}

\subsection{Proof of Theorem \ref{thm:rotation} (Delocalization via Rotation)}
\label{app:proof_hierarchy}

\textbf{Part 1: Random Rotation (Gaussian Limit).}
Let $\tilde{\mathbf{u}} = \mathbf{u} R$ where $R$ is a random orthogonal matrix. By the concentration of measure, elements of $\tilde{\mathbf{u}}$ converge to a Gaussian distribution $\tilde{u}_i \sim \mathcal{N}(0, \sigma^2)$ with $\sigma^2 = \| \mathbf{u} \|_2^2 / r$.
For a Gaussian variable $x$, $\mathbb{E}[|x|] = \sigma \sqrt{2/\pi}$. Thus, the expected $L_1$ norm is:
\begin{equation}
    \mathbb{E}[ \| \tilde{\mathbf{u}} \|_1 ] \approx r \cdot \sigma \sqrt{\frac{2}{\pi}} = \sqrt{\frac{2r}{\pi}} \| \mathbf{u} \|_2
\end{equation}
Substituting this into Lemma \ref{lemma:distortion_duality}:
\begin{equation}
    \mathbb{E}[\lambda_{\text{Rot}}] \approx 1 - \frac{1}{r} \left( \sqrt{\frac{2r}{\pi}} \right)^2 = 1 - \frac{2}{\pi} \approx 0.3634
\end{equation}
This proves that Random Rotation strictly bounds the distortion, reducing the error by $\sim$63.7\% compared to the worst-case SVD ($\lambda \approx 1$).

\textbf{Part 2: Joint-ITQ (Optimal Alignment).}
The Joint-ITQ algorithm maximizes the objective $\mathcal{J}(R) = \| \text{sign}(Z R) - Z R \|_F^2$ is minimized. This is equivalent to maximizing the trace:
\begin{equation}
    \max_R \text{Tr}(\text{sign}(Z R)^T (Z R)) = \max_R \sum_{i,j} \text{sign}((ZR)_{ij}) (ZR)_{ij} = \max_R \sum_{i,j} |(ZR)_{ij}| = \max_R \| Z R \|_1
\end{equation}
Thus, the ITQ optimization explicitly searches for a rotation $R$ that maximizes the $L_1$ norm of the latent factors. 
From Lemma \ref{lemma:distortion_duality}, $\lambda$ is monotonically decreasing with respect to the $L_1$ norm. Since ITQ starts from a random rotation and iteratively increases the $L_1$ norm, it guarantees a distortion coefficient strictly lower than Random Rotation:
\begin{equation}
    \lambda_{\text{ITQ}} \le \lambda_{\text{Rot}} < \lambda_{\text{SVD}}
\end{equation}
In the limit of perfect alignment (where vectors align with hypercube vertices), $\| Z R \|_1 = \sqrt{r} \| Z R \|_2$, yielding $\lambda_{\text{ITQ}} = 0$. \hfill $\blacksquare$

\begin{figure}[t]
\centering
\resizebox{0.60\columnwidth}{!}{
\begin{tikzpicture}[font=\sffamily]
    \begin{axis}[
        width=10cm, height=5cm,
        axis lines=left,
        axis line style={-Stealth, thick, icmlgray},
        xlabel={\textbf{Rank Index} ($k$)},
        ylabel={\textbf{Singular Value} ($\sigma_k^2$)},
        xtick=\empty, ytick=\empty,
        xmin=0, xmax=11,
        ymin=0, ymax=6.5,
        domain=0:11,
        samples=200,
        clip=false,
        xlabel style={at={(axis description cs:1.0,-0.1)}, anchor=north east, font=\scriptsize},
        ylabel style={at={(axis description cs:0,0.9)}, anchor=south west, font=\scriptsize, rotate=-90}%
    ]

        \def\curve{6/(\x+1)}
        
        \def\noisycurve{(3/(\x+1))} 

        \def\rfp{1.0}   
        \def\rbin{10.0}  

        \addplot[name path=top, domain=0:\rfp, draw=none] {\curve};
        \addplot[name path=bottom, domain=0:\rfp, draw=none] {\noisycurve};

        \addplot[
            fill=icmlblue!20!icmlgray!20, 
            opacity=0.7,
            domain=0:\rfp
        ] {\noisycurve} \closedcycle;

        \addplot[
            pattern={Lines[angle=45, distance={6pt}, line width={0.4pt}]},
            pattern color=icmlblue,
            draw=none, 
            opacity=0.8,
            domain=0:\rfp
        ] {\noisycurve} \closedcycle;

        \addplot[
            pattern={Lines[angle=-45, distance={6pt}, line width={0.4pt}]},
            pattern color=icmlgray,
            draw=none, 
            opacity=0.8,
            domain=0:\rfp
        ] {\noisycurve} \closedcycle;
        \addplot[icmlgray, opacity=0.6] fill between [of=top and bottom];

        \addplot[fill=icmlblue, opacity=0.7, domain=\rfp:\rbin] 
            {\noisycurve} \closedcycle;

        \addplot[thick, black!40, dashed, domain=0:\rbin] {\curve};

        \addplot[thick, black!80, domain=0:\rbin] {\noisycurve};

        \draw[dashed, icmlgray] (axis cs:\rfp, 0) -- (axis cs:\rfp, 3.0);
        \draw[dashed, icmlgray] (axis cs:\rbin, 0) -- (axis cs:\rbin, 6/11);

        \node[black!80, align=center, font=\bfseries\scriptsize] (loss) at (axis cs:3.5, 3.5) {Original $\sigma_k^2 \approx C^2k^{-2\gamma}$}; 
        \draw[->, black!80, thick] (loss.south) -- (axis cs:2.5, 2.0);
        \node[black!80, align=center, font=\bfseries\scriptsize] (gain) at (axis cs:7.0, 2.0) {Quantized $\propto (1-\Lambda)$};
        \draw[->, black!80, thick] (gain.south) -- (axis cs:6.5, 0.5);

        \node[anchor=north, font=\small] at (axis cs:\rfp, 0) {$r_{\text{FP16}}$};
        \node[anchor=north, font=\small] at (axis cs:\rbin, 0) {$r_{\text{Binary}}$};

        \node[draw=icmlgray!50, rounded corners, fill=white, anchor=north east, align=left, font=\scriptsize, drop shadow={opacity=0.15}, inner sep=6pt] 
            at (axis cs:11, 6.5) {
            \textbf{Spectral Break-Even}\\[0.3em]
            \textcolor{icmlgray}{\rule{6pt}{6pt}} \ \textbf{Strategy A:} Tiny Rank FP16 \\
            \textcolor{icmlblue}{\rule{6pt}{6pt}} \ \textbf{Strategy B:} Low Rank Binary \\[0.5em]
            \textit{For heavy-tailed weights ($\gamma < \gamma^*$):}\\
            \textbf{Tail Gain} $>$ \textbf{Quant. Cost}
        };
        
        \node[anchor=east, font=\scriptsize\bfseries, color=black!80, xshift=-2pt] at (axis cs:0, 6) {$\sigma_1^2$};
        \fill[black!40] (axis cs:0, 6) circle (1.5pt); 

        \node[anchor=east, font=\scriptsize\bfseries, color=black!80, xshift=-2pt] at (axis cs:0, 3) {$(1-\Lambda)\sigma_1^2$};
        \fill[black!80] (axis cs:0, 3) circle (1.5pt); 
        
        \draw[decorate, decoration={brace, amplitude=3pt}, thick, black!80] 
            (axis cs:-0.1, 3.1) -- (axis cs:-0.1, 5.9) 
            node[midway, left=4pt, font=\tiny, align=right] {$\Lambda\sigma_1^2$};
    \end{axis}
\end{tikzpicture}
}
\caption{\textbf{Conceptual Illustration of the Spectral Gain.} 
Comparison between Strategy A (\textcolor{icmlgray}{Tiny-Rank FP16}) and Strategy B (\textcolor{icmlblue}{Low-Rank Binary}) under a fixed budget. 
While Strategy A suffers from rank starvation, Strategy B utilizes massive rank expansion to capture the heavy-tailed energy.}
\label{fig:spectral_gap}
\vspace{-0.2in}
\end{figure}
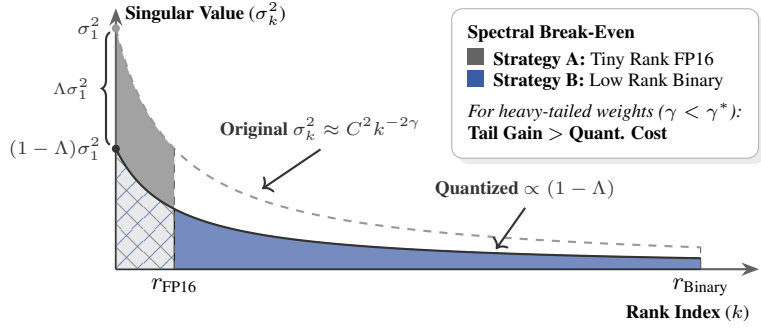

\section{Conceptual Visualization of the Spectral Gain}
\label{app:spectral_gap_viz}
In this section, we provide a visual interpretation of the theoretical trade-off discussed in Section \ref{sec:method}. Figure \ref{fig:spectral_gap} illustrates the \textit{Spectral Energy Gain} under a fixed memory budget. 

For heavy-tailed distributions characteristic of LLMs (where $\gamma < \gamma^*$), \textbf{Strategy A (Tiny-Rank FP16)} suffers from severe truncation error, discarding significant information residing in the tail. In contrast, \textbf{Strategy B (Low-Rank Binary)} tolerates quantization noise to achieve massive rank expansion. As visually demonstrated, the tail gain (information recovered by higher rank) significantly outweighs the quantization cost (noise area), validating the superiority of the sub-1-bit architecture.

\section{Detailed Architecture of LittleBit}
\label{app:littlebit_details}

In this section, we provide supplementary details on the initialization strategy and a comparative analysis of the LittleBit architecture, which were omitted from the main text.

\subsection{Dual-SVID Initialization}
\label{app:dual_svid}

Training a network with binary latent factors from scratch is notoriously unstable due to the non-differentiable nature of the sign function. Random initialization typically leads to training collapse. To address this, LittleBit employs \textbf{Dual-SVID (Sign-Value-Independent Decomposition)}, a specialized initialization scheme that decouples the initialization of binary geometry (Signs) and floating-point magnitudes (Values).

The Dual-SVID process operates in two main stages, assuming the weight matrix $\mathbf{W} \in \mathbb{R}^{d_{out} \times d_{in}}$:
\begin{enumerate}
    \item \textbf{Subspace Approximation via SVD:} 
    First, we approximate the pre-trained weight $\mathbf{W}$ using a standard singular value decomposition (SVD), truncated to the target rank $r$. We distribute the singular values equally to obtain the initialized latent factors $\hat{\mathbf{U}}$ and $\hat{\mathbf{V}}$ directly:
    \begin{equation}
        \mathbf{W} \approx \hat{\mathbf{U}} \hat{\mathbf{V}}^\top \quad \text{where} \quad \hat{\mathbf{U}} = U_r \Sigma_r^{1/2}, \ \hat{\mathbf{V}} = V_r \Sigma_r^{1/2}
    \end{equation}
    Here, $\hat{\mathbf{U}} \in \mathbb{R}^{d_{out} \times r}$ and $\hat{\mathbf{V}} \in \mathbb{R}^{d_{in} \times r}$. These factors preserve the optimal low-rank subspace direction and sign structure.

    \item \textbf{Scale Extraction via Rank-1 Decomposition:} 
    To initialize the floating-point scales ($\mathbf{h}, \mathbf{l}, \mathbf{g}$), we extract the magnitude envelopes from $\hat{\mathbf{U}}$ and $\hat{\mathbf{V}}$. Since the LittleBit architecture constraints the magnitude to a shared latent scale (Eq. \ref{eq:factorization}), we employ a \textbf{Rank-1 approximation} on the absolute values of these factors:
    \begin{equation}
        |\hat{\mathbf{U}}| \approx \mathbf{h} \cdot \boldsymbol{\ell}_{u}^\top, \quad |\hat{\mathbf{V}}| \approx \mathbf{g} \cdot \boldsymbol{\ell}_{v}^\top
    \end{equation}
    Here, $\mathbf{h} \in \mathbb{R}^{d_{out}}$ and $\mathbf{g} \in \mathbb{R}^{d_{in}}$ are directly assigned as the row and column scales, respectively. The vectors $\boldsymbol{\ell}_{u}, \boldsymbol{\ell}_{v} \in \mathbb{R}^r$ capture the relative importance of each latent dimension. The final central scale $\mathbf{l}$ is initialized as their element-wise product: $\mathbf{l} = \boldsymbol{\ell}_{u} \odot \boldsymbol{\ell}_{v}$.
\end{enumerate}

This strategy effectively disentangles multi-dimensional magnitude variations, ensuring that the initial effective weight $\hat{W}$ closely approximates the original $\mathbf{W}$ before quantization-aware training begins.

\begin{algorithm}[H]
\caption{Dual-SVID Initialization for LittleBit}
\label{alg:dual_svid}
\begin{algorithmic}[1]
\STATE \textbf{Input:} Pre-trained weight $W$, Target rank $r$
\STATE \textbf{Step 1: SVD Decomposition}
\STATE Compute $W \approx U \Sigma V^T$ via randomized SVD.
\STATE Truncate to rank $r$ to obtain $U_r, \Sigma_r, V_r$.
\STATE \textbf{Step 2: Latent Weight Initialization}
\STATE \textit{// Initialize full-precision latent weights for STE-based training}
\STATE Assign truncated SVD components to latent weights: $\hat{U} \leftarrow U_r \Sigma_{1:r}^{1/2}$, $\hat{V} \leftarrow V_r \Sigma_{1:r}^{1/2}$.
\STATE \textbf{Step 3: Magnitude Initialization (Scale)}
\STATE Approximate magnitudes $|\hat{U}_r|$ and $|\hat{V}_r|$ using Rank-1 approximations to initialize row/col scales $h, g, l$.
\STATE \textbf{Output:} Latent weights $\hat{U}, \hat{V}$ and Scale factors $h, g, l$.
\end{algorithmic}
\end{algorithm}

\section{Extended Results}
\label{app:extended_results}
\begin{table*}[h]
\caption{\textbf{Extended Results on Zero-shot Tasks.} Detailed breakdown of zero-shot performance across Llama-2 and Llama-3 families. We report Perplexity (PPL) on WikiText-2 and accuracy for HellaSwag (Hella), WinoGrande (Wino), ARC-Easy (ArcE), ARC-Challenge (ArcC), and PIQA. \textbf{Avg} denotes the average accuracy of the five zero-shot tasks.}
\label{tab:extended_results}
\vskip 0.15in
\begin{center}
\begin{small}
\begin{sc}
\resizebox{1.0\textwidth}{!}{
\setlength{\tabcolsep}{3.5pt} 
\begin{tabular}{l|c|ccccccc|ccccccc|ccccccc}
\toprule
\multirow{2}{*}{\textbf{Method}} & \multirow{2}{*}{\textbf{Bits}} & \multicolumn{7}{c|}{\textbf{Llama-3 8B}} & \multicolumn{7}{c|}{\textbf{Llama-2 7B}} & \multicolumn{7}{c}{\textbf{Llama-2 13B}} \\
\cmidrule(lr){3-9} \cmidrule(lr){10-16} \cmidrule(lr){17-23}
 & & \textbf{PPL}$\downarrow$ & Wino$\uparrow$ & Hella$\uparrow$ & ArcE$\uparrow$ & ArcC$\uparrow$ & PiQA$\uparrow$ & \textbf{Avg}$\uparrow$ & \textbf{PPL}$\downarrow$ & Wino$\uparrow$ & Hella$\uparrow$ & ArcE$\uparrow$ & ArcC$\uparrow$ & PiQA$\uparrow$ & \textbf{Avg}$\uparrow$ & \textbf{PPL}$\downarrow$ & Wino$\uparrow$ & Hella$\uparrow$ & ArcE$\uparrow$ & ArcC$\uparrow$ & PiQA$\uparrow$ & \textbf{Avg}$\uparrow$ \\
\midrule
\textit{FP16} & 16 & 6.10 & 73.24 & 79.09 & 80.01 & 53.16 & 79.33 & 72.97 & 5.47 & 68.59 & 75.93 & 76.18 & 46.08 & 77.64 & 68.88 & 4.88 & 72.30 & 79.36 & 79.50 & 48.98 & 79.22 & 71.87 \\
\midrule
GPTQ & 2 & 1480 &49.33 &27.36 &27.40 &22.87 &53.08 & 36.01 & 52.2 &49.80 &34.20 &29.55 &24.57 &55.82 & 38.79 & 23.6 &50.59 &44.03 &40.11 &25.85 &61.10 & 44.34 \\
EfficientQAT & 2 & 9.80 &65.67 &67.78 &68.98 &39.93 & 75.08& 63.49 & 7.17 &64.17 &67.52 &71.04 &38.05 &73.89 & 62.93 & 6.08 &69.22 &74.43 &75.04 &44.80 &76.82 & 68.06 \\
\midrule
\multicolumn{23}{c}{\textit{1-bit Regime}} \\
\midrule
BiLLM & 1.1 & 59.37 & 49.27 & 30.44 & 32.11 & 20.99 & 55.82 & 37.73 & 29.00 & 51.14 & 33.76 & 36.57 & 23.12 & 59.47 & 40.81 & 21.53 & 57.22 & 44.71 & 51.81 & 27.39 & 64.47 & 49.12 \\
ARB-LLM & 1.1 & 27.63 & \textbf{61.09}& 47.76 & 57.70 & 29.86 & 67.36 & 52.75 & 15.88 & \textbf{60.93} & 49.69 & 54.80 & \textbf{29.35} & 66.81 & 52.32 & 12.13 & \textbf{65.82} & 56.59 & 63.85 & 33.36 & 70.29 & 57.98 \\
OneBit & 1.0 & 13.09 & 52.88 & 53.28 & 56.27 & 28.33 & 70.40 & 52.23 & 8.36 & 56.91 & 57.70 & \textbf{59.01} & 28.24 & 71.76 & 54.72 & 7.41 & 60.22 & 63.67 & 63.09 & 31.91 & 73.45 & 58.47 \\
LittleBit & 1.0 & 16.30 & 49.17 & 42.77 & 49.24 & 25.26 & 69.10 & 47.11 & 9.08 & 56.35 & 54.40 & 54.80 & 27.73 & 70.78 & 52.81 & 8.18 & 54.78 & 52.06 & 53.11 & 27.13 & 70.35 & 51.49 \\
\textbf{LittleBit-2} & 1.0 & \textbf{11.53} & 60.85 & \textbf{60.22} & \textbf{61.20} & \textbf{30.72} & \textbf{73.67} & \textbf{57.33} & \textbf{8.27} & 58.88 & \textbf{59.48} & 58.50 & 29.27 & \textbf{72.80} & \textbf{55.79} & \textbf{7.37} & 63.46 & \textbf{65.74} & \textbf{65.57} & \textbf{33.79} & \textbf{74.10} & \textbf{60.53} \\
\midrule
\multicolumn{23}{c}{\textit{Sub-1-bit Regime}} \\
\midrule

STBLLM & 0.55 & 241.96 & 48.15 & 28.71 & 29.59 & 23.12 & 54.24 & 36.76 & 32.93 & 52.80 & 37.77 & 40.78 & 24.49 & 61.15 & 43.40 & 27.05 & 57.38 & 38.75 & 45.50 & 24.49 & 61.97 & 45.62 \\
LittleBit & 0.55 & 18.47 & 51.30 & 36.30 & 46.76 & 22.70 & 65.89 & 44.59 & 10.47 & 52.96 & 46.20 & 49.54 & 25.09 & 68.66 & 48.49 & 9.24 & 55.64 & 53.18 & 53.28 & 26.96 & 69.75 & 51.76 \\
\textbf{LittleBit-2} & 0.55 & \textbf{14.01} & \textbf{55.96} & \textbf{50.71} & \textbf{54.21} & \textbf{26.02} & \textbf{69.75} & \textbf{51.33} & \textbf{9.65} & \textbf{55.49} & \textbf{51.05} & \textbf{52.19} & \textbf{27.22} & \textbf{68.88} & \textbf{50.97} & \textbf{8.56} & \textbf{59.43} & \textbf{57.35} & \textbf{59.68} & \textbf{29.61} & \textbf{72.14} & \textbf{55.64} \\
\cmidrule{1-23}
LittleBit & 0.1 & 26.11 & \textbf{51.70} & 30.02 & 40.32 & \textbf{23.46} & \textbf{63.49} & \textbf{41.80} & 15.92 & 51.22 & 30.55 & 39.02 & \textbf{24.40} & 61.48 & 41.33 & 15.09 & \textbf{51.14} & 31.18 & 38.13 & 21.93 & 62.57 & 40.99 \\
\textbf{LittleBit-2} & 0.1 & \textbf{23.74} & 50.67 & \textbf{30.66} & \textbf{41.04} & 22.53 & 63.28 & 41.64 & \textbf{14.70} & \textbf{52.57} & \textbf{32.61} & \textbf{39.73} & 22.27 & \textbf{63.11} & \textbf{42.06} & \textbf{13.94} & 50.36 & \textbf{33.24} & \textbf{40.87} & \textbf{23.72} & \textbf{62.73} & \textbf{42.18} \\
\bottomrule
\end{tabular}
}
\end{sc}
\end{small}
\end{center}
\vskip -0.1in
\end{table*}

\newpage

\section{$\gamma$ Validation}
\label{app:gamma_val}

\begin{figure}[h]
    \centering
    \includegraphics[width=1.0\linewidth]{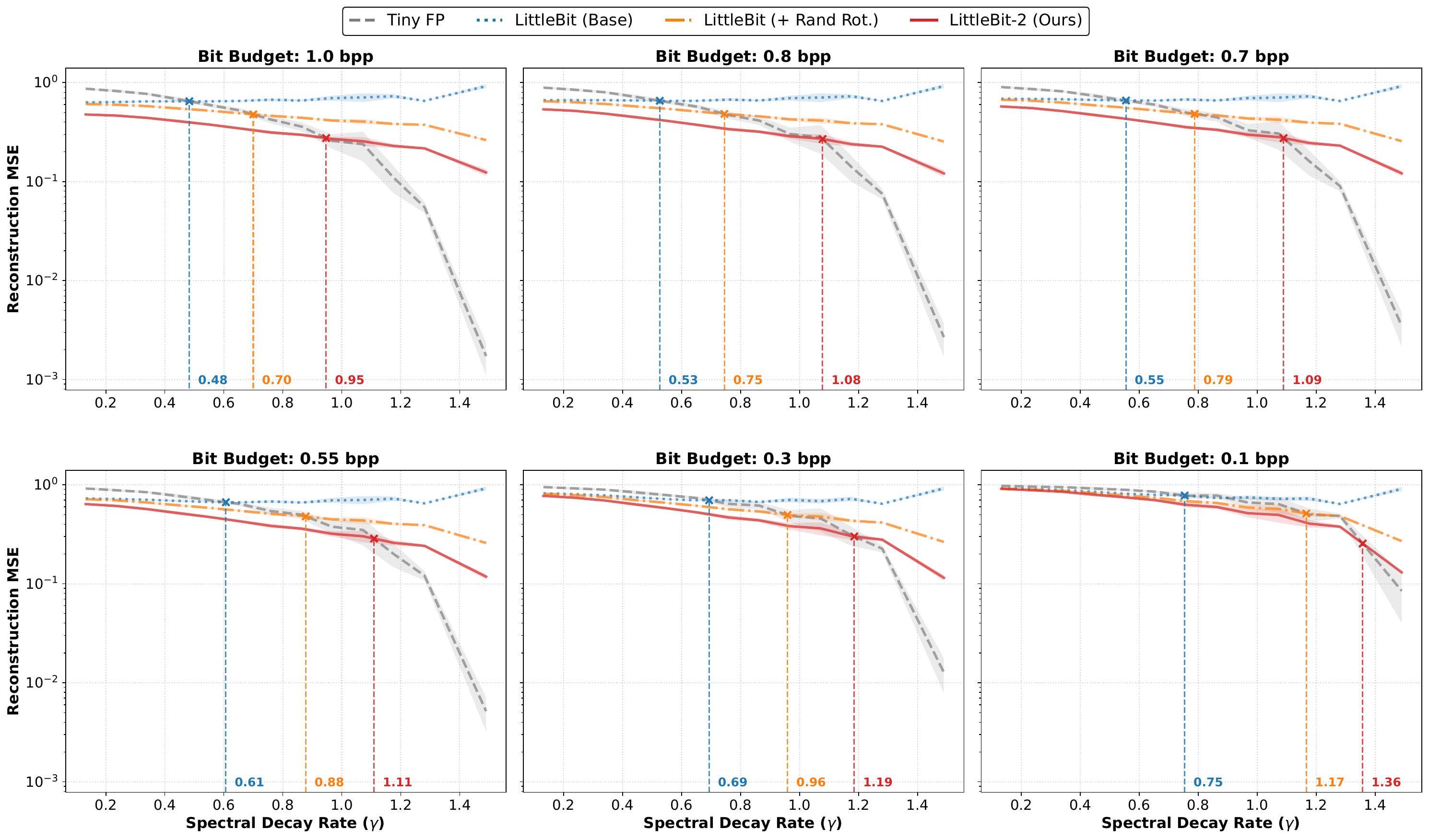}
    \caption{\textbf{Empirical Spectral Break-Even Analysis.} 
    Reconstruction error (MSE) trends across varying $\gamma$ values for bit-rates from 1.0 down to 0.1 bpp. The lines represent the average error calculated from the actual weights of 8 distinct LLMs. 
    Consistent with the synthetic validation, \textbf{LittleBit-2} (Red) dominates the heavy-tailed regime ($\gamma \lesssim 0.5$) and significantly extends the superiority range over FP16 (Gray) compared to the baseline LittleBit (Blue), especially in extreme compression regimes.}
    \label{fig:gamma_trend_analysis}
\end{figure}
To validate the theoretical bounds derived in Section \ref{sec:why_binary} under realistic conditions, we performed a comprehensive trend analysis using actual weight matrices from 8 different models, including the Llama-2, Llama-3, and Gemma-3 families. We measured the initialization reconstruction error (MSE) across varying spectral decay rates ($\gamma$) for bit-budgets ranging from 1.0 down to 0.1 bpp. 
Specifically, we aggregated the MSE of all linear layers from the 8 models to observe the performance crossover points. As predicted, in the heavy-tailed regime (lower $\gamma$), the Tiny-Rank FP16 baseline suffers from rank starvation, whereas LittleBit-2 consistently achieves the lowest error. Notably, as the compression becomes more extreme (lower bpp), the crossing point shifts, indicating that our geometric alignment strategy becomes increasingly critical for preserving information in heavy-tailed distributions.

\begin{figure}[h]
    \centering
    \includegraphics[width=0.5\linewidth]{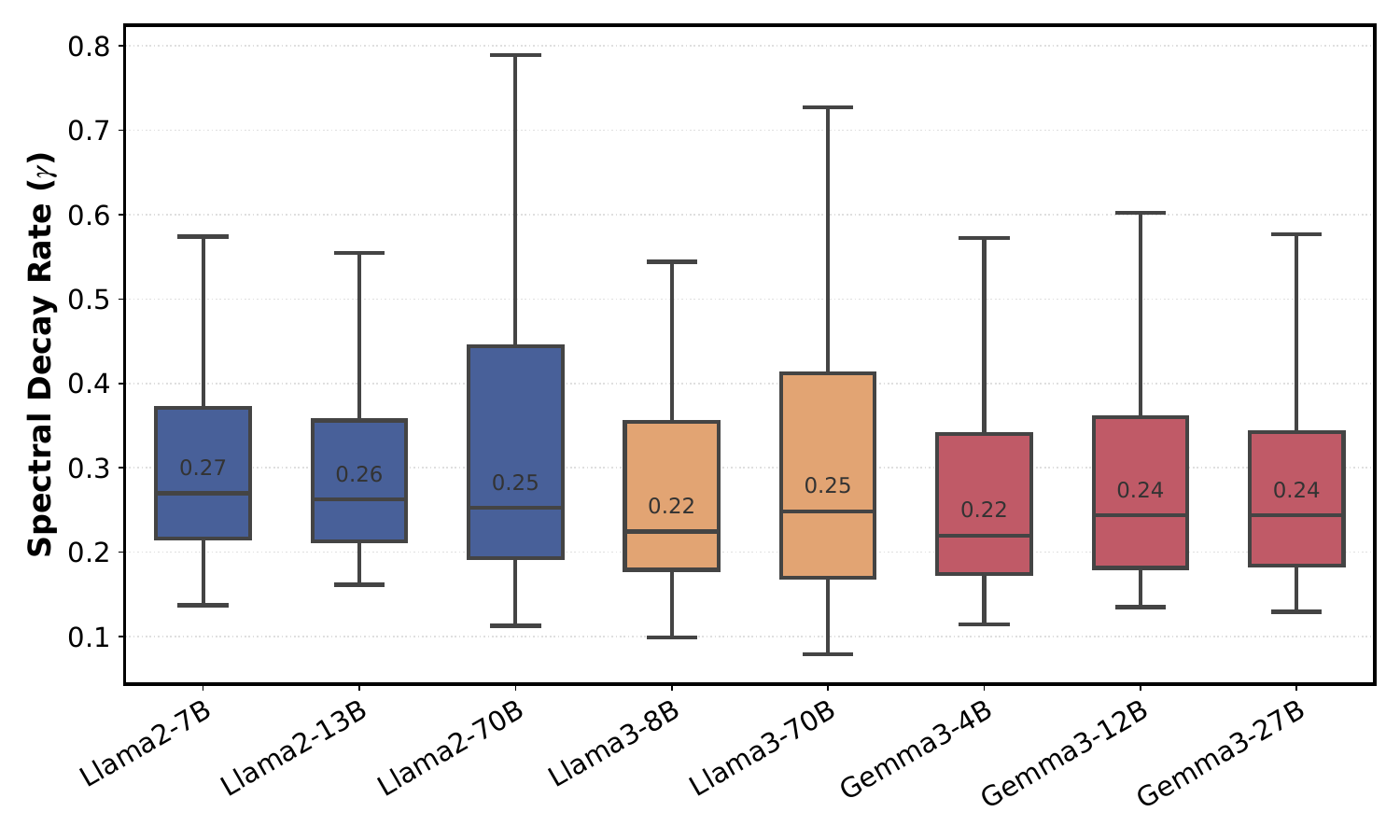}
    \caption{\textbf{Distribution of Spectral Decay Rates ($\gamma$) by Model.} Box plots summarizing the $\gamma$ distribution of all linear layers across 8 different models. The median values consistently fall within the heavy-tailed range of $[0.26, 0.33]$.}
    \label{fig:model_gamma_dist}
\end{figure}

\begin{figure}[h]
    \centering
    \includegraphics[width=0.5\linewidth]{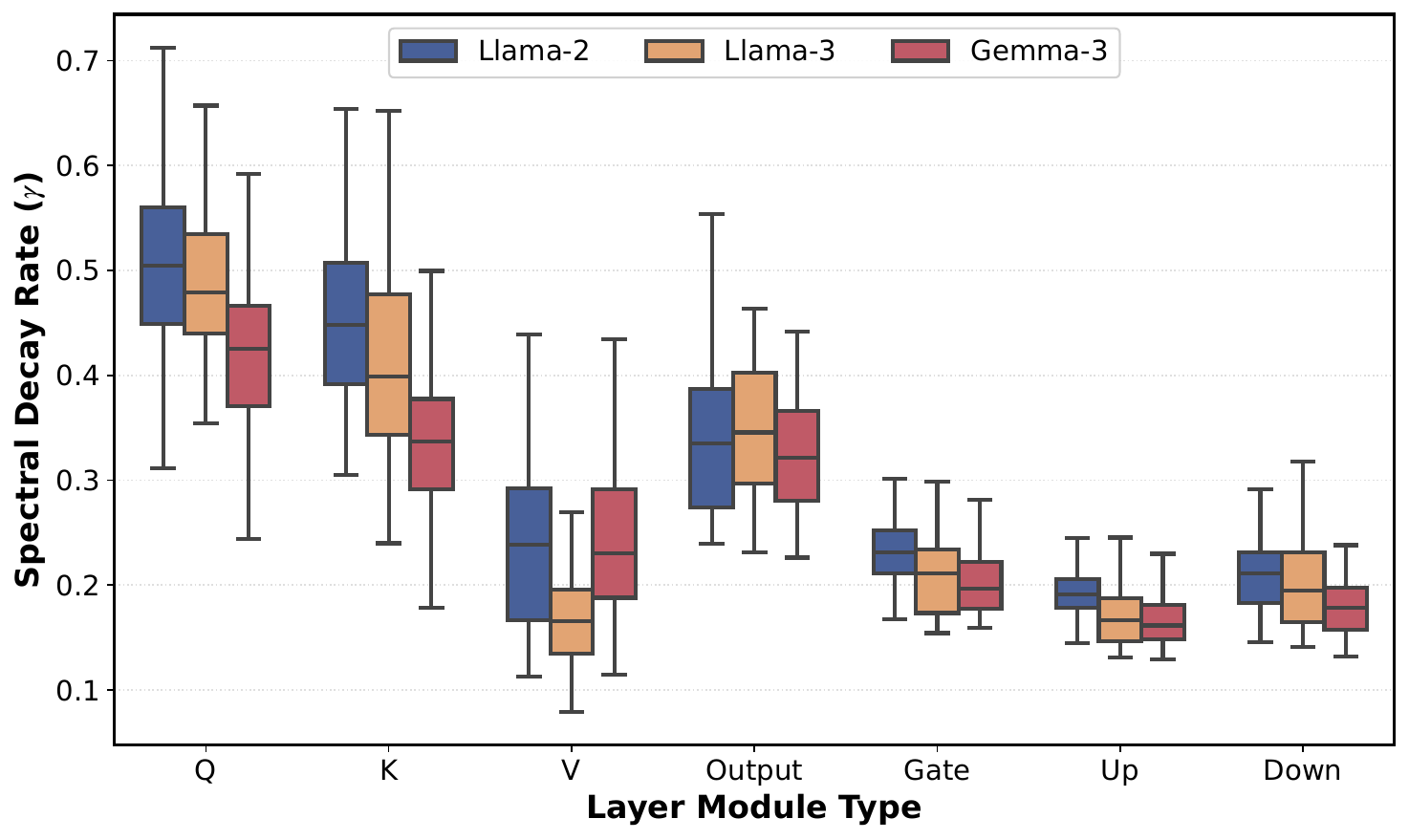}
    \caption{\textbf{Distribution of $\gamma$ by Module Type.} Analysis of spectral decay rates grouped by projection types (Query, Key, Value, Output, etc.) across the Llama-2, Llama-3, and Gemma-3 families.}
    \label{fig:module_gamma_dist}
\end{figure}

\newpage

\section{Extended Ablation Study}
\label{app:ext_ablation}
\subsection{Sensitivity to Joint-ITQ Iterations}
\label{subsec:itq_sensitivity}

The \textbf{Joint-ITQ} algorithm iterates to optimally align latent factors with the binary hypercube. A critical hyperparameter is the number of iterations $T$, which governs the trade-off between the geometric alignment quality (reconstruction error) and the initialization overhead.

To determine the optimal $T$, we investigated the impact of iteration counts on both the Mean Squared Error (MSE) and the wall-clock initialization time. We performed this ablation on the query projection layer (\texttt{q\_proj}) of the 15th block in Llama-2 7B, sweeping $T$ from 0 to 100.

\begin{figure}[h]
    \centering
    \includegraphics[width=0.50\linewidth]{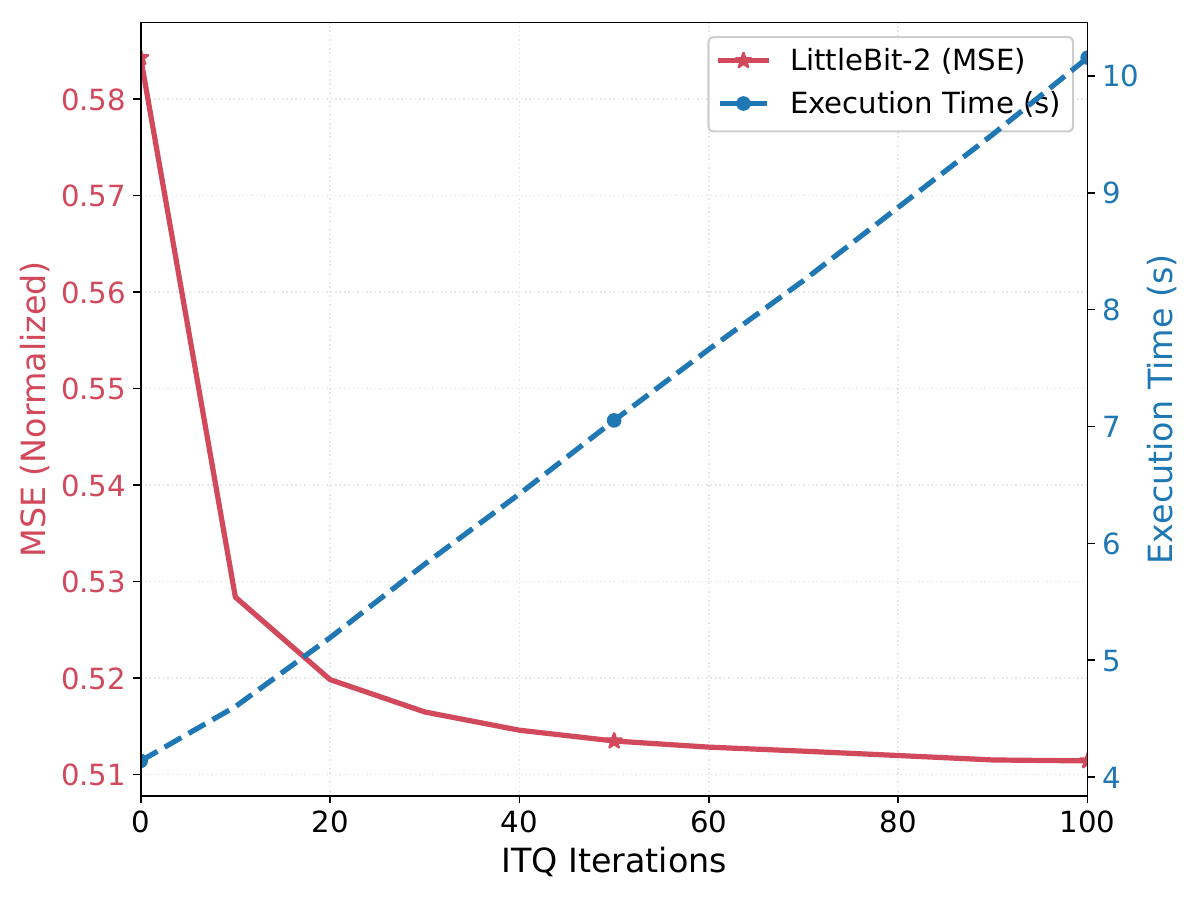}
    \caption{\textbf{Convergence vs. Overhead Analysis (Joint-ITQ).} 
    Dual-axis plot showing the evolution of reconstruction error and initialization time.
    \textcolor{icmlred}{\textbf{Left Axis (Red):}} The MSE exhibits a steep descent in the early phase and saturates around $T=50$, indicating that the latent geometry rapidly converges to the optimal bimodal orientation.
    \textcolor{icmlblue}{\textbf{Right Axis (Blue):}} The initialization time scales linearly. The overhead for 50 iterations is marginal ($\sim$3s) compared to the baseline ($T=0$). We adopt $T=50$ as the optimal operating point.}
    \label{fig:iter_ablation}
\end{figure}

\paragraph{Convergence vs. Overhead.}
Figure \ref{fig:iter_ablation} summarize the trajectory of error and time.
\begin{itemize}
    \item \textbf{Reconstruction Fidelity (\textcolor{icmlred}{Red}):} The MSE exhibits a sharp decay within the first 20 iterations and reaches a saturation point of diminishing returns near $T=50$. This confirms that the joint optimization efficiently identifies the principal geometric alignment.
    \item \textbf{Computational Cost (\textcolor{icmlblue}{Blue}):} The initialization time increases linearly but remains computationally inexpensive. The baseline Random Rotation ($T=0$) requires approximately 4 seconds due to SVD and allocation. Extending the optimization to $T=50$ incurs a modest overhead of $\sim$3 seconds (Total $\sim$7 seconds).
\end{itemize}

Given that this process is a \textbf{one-time offline initialization} prior to QAT and incurs \textbf{zero inference overhead}, the additional latency is a highly cost-effective investment for the significant reduction in quantization noise. Consequently, we fix $T=50$ for all main experiments to ensure robust convergence.

\section{Efficacy of Residual}
\label{app:residual}
To maximize representational fidelity under a strict bit budget, we employ a residual architecture where the approximation is split into a primary path and a residual path ($W \approx \hat{W}_{\text{pri}} + \hat{W}_{\text{res}}$). In this section, we provide the theoretical justification for this design and empirically validate its impact.

A fundamental question arises: \textit{Does splitting the rank into two paths offer a structural advantage over a single, wider path?} The answer depends on the precision domain.

\begin{figure}[h]
    \centering
    \includegraphics[width=0.50\linewidth]{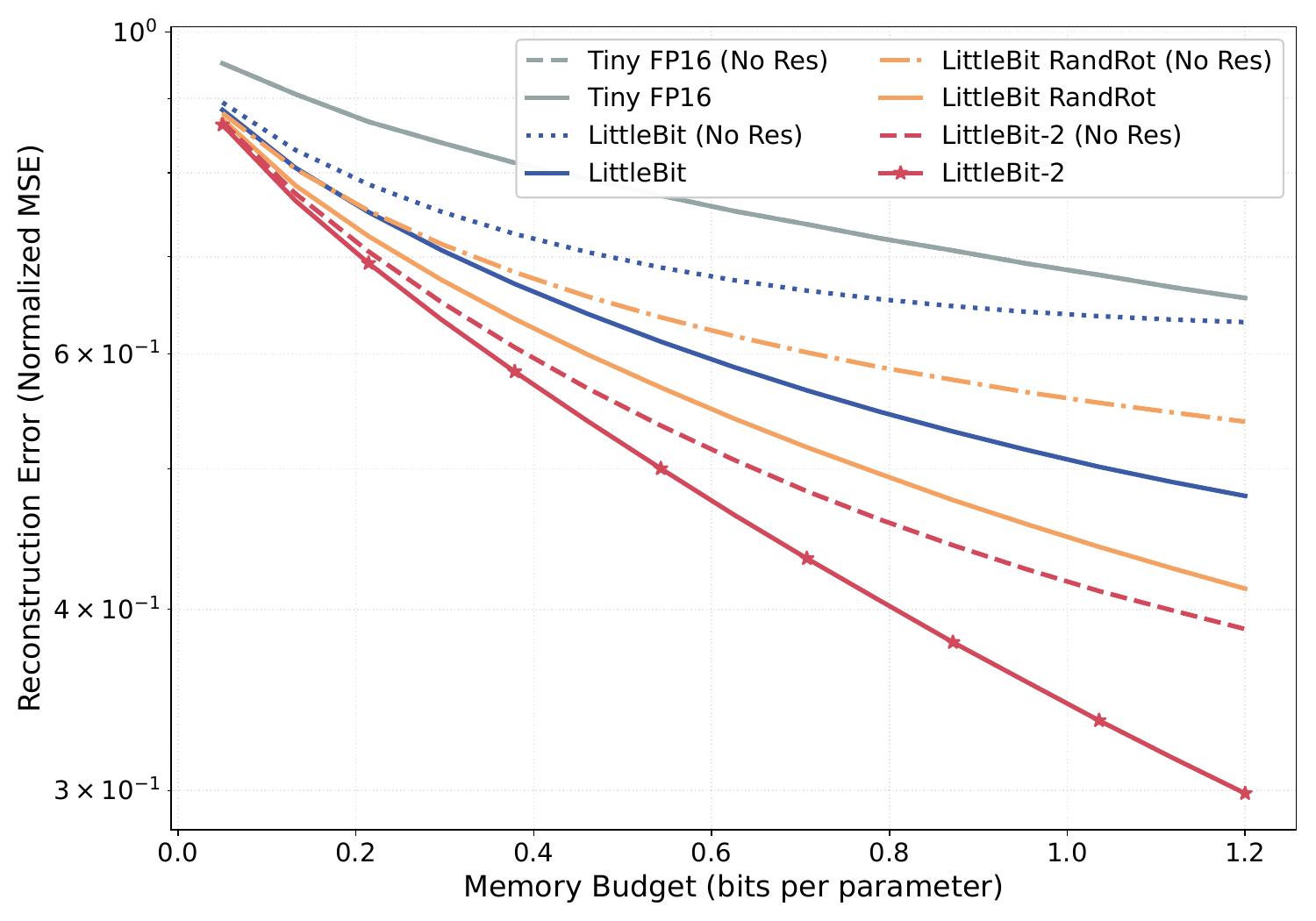}
    \caption{\textbf{Impact of Residual Architecture on Reconstruction Error.} 
    Comparison of Mean Squared Error (MSE) versus Memory Budget. Solid lines indicate the residual architecture, and dashed lines indicate the single-path ablation.
    The results confirm a strict performance hierarchy:
    (1) \textbf{FP16 Invariance:} For Tiny-Rank FP16, the residual and non-residual lines overlap, confirming that splitting rank offers no benefit in the linear regime.
    (2) \textbf{Residual Gain:} For all binary methods (LittleBit, LittleBit(+Rot.), LittleBit-2, the residual architecture (Solid) consistently achieves lower MSE than the non-residual counterparts (Dashed).
    (3) \textbf{Geometric Dominance:} Notably, \textbf{LittleBit-2 (No Res)} outperforms the residual versions of prior methods (e.g., LittleBit(+Rot.)). This implies that efficacy of optimized geometric alignment (Joint-ITQ) is further maximized by the residual path.}
    \label{fig:residual_efficacy}
\end{figure}

\textbf{Linear Regime (FP16).} Due to the linearity of matrix factorization, decomposing a rank-$2r$ approximation into two rank-$r$ matrices ($U_1 V_1^T + U_2 V_2^T$) is mathematically equivalent to a single rank-$2r$ SVD. Thus, for the \textbf{Tiny-Rank FP16} baseline, the residual topology offers no information-theoretic gain.

\textbf{Non-Linear Regime (Binary).} In the binary domain, the quantization operator $\mathcal{Q}(\cdot)$ introduces non-linear noise $E = W - \mathcal{Q}(W_{\text{pri}})$. A single binary path is strictly bound by this noise floor. However, in a residual architecture, the secondary path $\hat{W}_{\text{res}}$ explicitly targets the approximation of this error term $E$. This acts as a \textbf{Coarse-to-Fine Refinement} mechanism, where the primary path captures the structural skeleton and the residual path corrects the quantization noise.

We validate this hypothesis by analyzing the reconstruction error (MSE) across varying memory budgets (0.05 $\sim$ 1.2 bpp). Figure \ref{fig:residual_efficacy} compares the performance of single-path (``No Res'') and residual architectures across different methods.

As illustrated in Figure \ref{fig:residual_efficacy}, the methods exhibit a clear hierarchy in reconstruction fidelity. The error decreases in the following order:
$$ \text{FP16} \approx \text{FP16(NoRes)} > \text{LittleBit} > \text{RandRot} > \textbf{LittleBit-2(NoRes)} > \textbf{LittleBit-2} $$
This empirical evidence supports two key conclusions: First, residual connections are essential for binary quantization to mitigate noise. Second, the geometric alignment provided by LittleBit-2 acts as a fundamental performance multiplier, establishing a new state-of-the-art even without residual enhancement.

\section{Memory Requirement Analysis}
\label{app:memory_analysis}

In this section, we explicitly define the memory requirement calculations used to evaluate the effective bits-per-parameter (bpp) and total model size. We denote the linear layer input dimension as $d_{in}$ and output dimension as $d_{out}$. The total parameter count is $N = d_{in} \cdot d_{out}$. Unless otherwise stated, all high-precision scaling factors are calculated in FP16 (16-bit).

\subsection{Standard Quantization Baselines}

\paragraph{GPTQ / EfficientQAT.}
For 2-bit quantization methods utilizing group-wise scaling, such as GPTQ \cite{frantar2022gptq} and EfficientQAT \cite{chen2025efficientqat}, we use a block size of $k=128$. Each block requires one FP16 scale and one FP16 zero-point.
\begin{equation}
    \mathcal{M}_{\text{GPTQ}} = \underbrace{2 \cdot N}_{\text{Quantized Weights}} + \underbrace{\frac{N}{128} \cdot (16 + 16)}_{\text{Scales \& Zero-points}} = 2.25 \cdot N \quad (\text{bits})
\end{equation}

\paragraph{OneBit.}
OneBit \cite{xu2024onebit} decomposes the weight matrix into a 1-bit matrix and two high-precision scaling vectors (row and column scales).
\begin{equation}
    \mathcal{M}_{\text{OneBit}} = \underbrace{1 \cdot N}_{\text{Binary Weights}} + \underbrace{16 \cdot (d_{in} + d_{out})}_{\text{Row \& Col Scales}}
\end{equation}

\subsection{Salient-Weight Binarization (BiLLM \& ARB-LLM)}
For BiLLM and ARB-LLM, we strictly follow the memory computation formulas presented in the ARB-LLM supplementary material \cite{li2024arb}. These methods employ a block-wise structure ($k=128$) and retain a subset of salient columns $c$ in higher precision. To align with their notation, we map $n = d_{out}$ and $m = d_{in}$.

\paragraph{BiLLM.}
Based on Equations 134 and 135 in the ARB-LLM supplementary material, the memory footprint includes the salient weights (2-bit), the binary base (1-bit), two levels of block scaling factors, and metadata bitmaps.
\begin{equation}
\begin{aligned}
    \mathcal{M}_{\text{BiLLM}} = & \underbrace{2 \cdot n \cdot c + \left\lceil \frac{m}{k} \right\rceil \cdot 3n \cdot 16}_{\text{Second-order Binarization}} \\
    & + \underbrace{n \cdot (m - c) + \left\lceil \frac{m}{k} \right\rceil \cdot 2n \cdot 16 \cdot 2}_{\text{First-order Binarization}} + \underbrace{{n \cdot m} + {m}}_{\text{Bitmaps}},
\end{aligned}
\end{equation}
where $c$ denotes the number of salient columns. In our analysis, we set $c = 128$.

\paragraph{ARB-LLM (RC).}
We utilize the ARB-RC (Row-Column) variant. Based on Equations 140 and 142 in the supplementary material, the memory is calculated as:
\begin{equation}
\begin{aligned}
    \mathcal{M}_{\text{ARB}} = & \underbrace{2 \cdot n \cdot c + \left(\left\lceil \frac{m}{k} \right\rceil \cdot 2n + 2c\right) \cdot 16}_{\text{Second-order Binarization}} \\
    & + \underbrace{n \cdot (m - c) + \left(\left\lceil \frac{m}{k} \right\rceil \cdot n + (m - c)\right) \cdot 16 \cdot 2}_{\text{First-order Binarization}} + \underbrace{{n \cdot m} + m}_{\text{Bitmaps}}
\end{aligned}
\end{equation}
In our experiments, we set the number of salient columns $c$ as 128.

\subsection{LittleBit Framework (LittleBit-1 \& LittleBit-2)}
The memory requirements for LittleBit-1 and LittleBit-2 are identical, governed by the \textbf{Residual Low-Rank Binary Factorization}. We utilize a residual architecture ($path=2$). The total memory bits $\mathcal{M}_{\text{LittleBit}}$ is calculated as:
\begin{equation}
    \mathcal{M}_{\text{LittleBit}} = \underbrace{2 r (d_{in} + d_{out} + 16)}_{\text{Rank-Dependent Terms}} + \underbrace{32 (d_{in} + d_{out})}_{\text{Fixed I/O Scales}}
\end{equation}
where $r$ is the latent rank. The components are broken down as follows:
\begin{itemize}
    \item \textbf{Binary Factors:} $2r(d_{in} + d_{out})$ bits. Two sets of binary matrices $U_b, V_b$ for the main and residual paths.
    \item \textbf{I/O Scales:} $32(d_{in} + d_{out})$ bits. FP16 Input/Output vectors ($g, h$) for both paths ($2 \text{ paths} \times 16 \text{ bits} = 32$).
    \item \textbf{Latent Scales:} $32r$ bits. This term (derived from $2r \cdot 16$) represents the latent scaling overhead. In our optimized inference, the two scalar vectors ($l_u, l_v$) per path are merged into a single FP16 vector $l \in \mathbb{R}^r$ per path ($2 \text{ paths} \times r \times 16 \text{ bits} = 32r$).
\end{itemize}

This formula allows us to precisely determine the maximum rank $r$ for any target bit-budget $\mathcal{B}$ by inverting the equation:
\begin{equation}
    r = \left\lfloor \frac{\mathcal{B} \cdot N - 32(d_{in} + d_{out})}{2(d_{in} + d_{out} + 16)} \right\rfloor
\end{equation}
Unlike the sparse baselines, LittleBit requires \textbf{no additional metadata indices or bitmaps}, ensuring that the calculated memory budget translates directly to the model size.

\paragraph{Model-Level Aggregation.}
To determine the total memory footprint reported in the experimental results (Section~\ref{sec:experiments}), we apply the aforementioned formulas to each linear layer within the model. We calculate the required bits based on the specific dimensions ($d_{in}, d_{out}$) of the Query, Key, Value, Output, Gate, Up, and Down projection layers. For all other parameters (e.g., LayerNorm, Embeddings), we assume standard FP16 precision. These per-layer memory costs are then aggregated across all transformer blocks defined by the model architecture. We report both the size of the model body (excluding the LM head) and the total model size (including the LM head) to provide a comprehensive comparison.

\section{Generated Samples}
\label{app:generate_samples}

We present qualitative generation samples from \textbf{Llama-2 13B} across different quantization regimes. To evaluate the preservation of knowledge, we use the fixed prompt: \textit{``Turing is called the father of computer science. Computer science is''}.

\newcommand{\gensample}[2]{%
    \begin{center}
    \fcolorbox{icmlgray}{white}{%
    \begin{minipage}{0.95\linewidth}
        \vspace{0.05in}
        \small
        \textbf{#1} \\
        \rule{\linewidth}{0.4pt} \\[-0.1in]
        \begin{description}
            \item[\textmd{\textit{Output:}}] #2
        \end{description}
    \end{minipage}
    }
    \end{center}
    \vspace{-0.1in}
}

\subsection{0.1-bit Regime (0.1 bpp)}

\gensample{LittleBit (0.1 bpp)}{
\underline{Turing is called the father of computer science. Computer science is} a branch of mathematics, and it's the science that gives computers the ability to perform calculations on things they want. It doesn't have much in common with logic or programming.
But what about those who don't know how to do math? That's exactly what I'm doing here: I'll teach you all these things you can remember from the ground up. You can even use them in your own programs!
There are 24 different types of tutoring programs offered by The University of Michigan. There are online tutorials available for free (if you don't already have one), there are
}

\gensample{LittleBit-2 (Ours) (0.1 bpp)}{%
\underline{Turing is called the father of computer science. Computer science is} a discipline that has been around for more than 10,000 years. Today it is one of the most important fields in the world and also one of the fastest growing areas of research.
The term ``Turing'' refers to the theory of time which has been developed over many centuries by scientists from different parts of the globe who have applied this knowledge to create computers. The history of computers began with the invention of the steam engine in 1937 but after that, the technology became advanced enough so that the first person to use it was an engineer named John Wiley. He invented
}

\subsection{1-bit Regime (1.0 bpp)}

\gensample{LittleBit (1.0 bpp)}{%
\underline{Turing is called the father of computer science. Computer science is} a sub-discipline within computer science and encompasses many fields, including artificial intelligence, software engineering, data structures and algorithms, computer architecture, algorithmics, computational complexity, and software development processes. The field also includes mathematical analysis and theoretical computer science.
The first computer was developed by Alan Turing in 1946 and the world has never looked back since then. In his book “Alan Turing: A biography”, Sir John Simon wrote about Turing’s fascinating life and achievements. He had a very unique way of thinking and he was an extremely creative person who could
}

\gensample{OneBit (Reference) (1.0 bpp)}{%
\underline{Turing is called the father of computer science. Computer science is} a vastly important subject that many people are not very familiar with. The reason for this is because it is an extremely complicated field that has so much to offer in its various aspects. In fact, there are several other fields like artificial intelligence and machine learning as well.
The most important thing about turing is that we can use it to build machines that actually do things similar to us humans. This is what makes it such an interesting concept. It is used to help solve some real-world problems that are difficult or impossible for human beings to figure out.
This article will discuss how turing works, why it matters and
}

\gensample{LittleBit-2 (Ours) (1.0 bpp)}{%
\underline{Turing is called the father of computer science. Computer science is} a branch of science that studies the feasibility and potentiality of computers, their construction, operation, application and impact on society.
In 1936 he proposed his theory of computation. In 1947 he developed the Turing test which was used to determine whether or not a machine could think for itself. It has since been widely adopted by researchers as an indicator of intelligence. He also contributed significantly to cryptography and computer security.
Alan Turing was born in 1900 in London England. His family were Jews who had emigrated from Russia at the turn of the century.
}

\section{Codes of Initialization Strategies}
\label{app:codes}

In this section, we provide the reference implementation for the three initialization strategies discussed in Section \ref{sec:method}: (1) Standard LittleBit, (2) Random Rotation, and (3) LittleBit-2 with Joint-ITQ. We abstract the core logic into a \texttt{LatentInitializer} class.

\subsection{Rank-1 Scale Extraction}
We first define the helper function to approximate the floating-point scales ($h, l, g$) from the latent factor magnitudes.

\begin{lstlisting}[language=Python, caption={Rank-1 Decomposition for Scale Extraction}]
def rank_one_decompose(X):
    """
    Approximates magnitude matrix X (|U| or |V|) via Rank-1 SVD.
    Returns vectors u_vec, v_vec such that X ~ u_vec @ v_vec
    """
    # 1. SVD on magnitudes
    U, S, Vh = torch.linalg.svd(X, full_matrices=False)
    
    # 2. Distribute the dominant singular value
    sqrt_s0 = torch.sqrt(S[0])
    u_vec = (U[:, :1] * sqrt_s0).contiguous()  # Shape: (N, 1)
    v_vec = (sqrt_s0 * Vh[:1, :]).contiguous() # Shape: (1, M)
    
    return u_vec, v_vec
\end{lstlisting}

\subsection{Initialization Algorithms}
The following methods implement the progression from the baseline to our proposed method.

\begin{lstlisting}[language=Python, caption={Initialization Strategies: Baseline, Rotation, and Joint-ITQ}]
class LatentInitializer:
    def __init__(self, rank):
        self.rank = rank

    def _extract_scales(self, U, V):
        """Extracts Tri-Scales (h, l, g) from latent factors."""
        # Decompose magnitudes: |U| -> h, l_u  AND  |V| -> l_v, g
        h, l_u = rank_one_decompose(torch.abs(U))
        l_v, g = rank_one_decompose(torch.abs(V))
        
        # Merge central scales
        l = (l_u.view(-1) * l_v.view(-1)).view(1, self.rank)
        return h, l, g

    def init_standard(self, W):
        """
        LittleBit Initialization.
        """
        # 1. Truncated SVD
        U_f, S_f, Vh_f = torch.linalg.svd(W.float(), full_matrices=False)
        
        # 2. Symmetric Scaling
        S_sqrt = torch.diag(torch.sqrt(S_f[:self.rank]))
        U = U_f[:, :self.rank] @ S_sqrt
        V = S_sqrt @ Vh_f[:self.rank, :]
        
        # 3. Extract Scales
        h, l, g = self._extract_scales(U, V)
        return U, V, h, l, g

    def init_random_rotation(self, W):
        """
        Internal Random Rotation.
        """
        # 1. Get Base Factors
        U, V, _, _, _ = self.init_standard(W)
        
        # 2. Apply Random Orthogonal Rotation
        # W ~ (U @ R) @ (R.T @ V)
        R = torch.empty((self.rank, self.rank), device=W.device)
        torch.nn.init.orthogonal_(R)
        
        U_rot = U @ R
        V_rot = R.t() @ V
        
        # 3. Extract Scales from Rotated Factors
        h, l, g = self._extract_scales(U_rot, V_rot)
        return U_rot, V_rot, h, l, g

    def init_joint_itq(self, W, n_iter=50):
        """
        LittleBit-2: Joint-ITQ Alignment.
        """
        # 1. Get Base Factors
        U, V, _, _, _ = self.init_standard(W)
        
        # 2. Construct Joint Manifold Z = [U; V.T]
        # Align input (V.T) and output (U) spaces simultaneously
        Z = torch.cat([U, V.t()], dim=0) 
        
        # 3. Solve Joint Orthogonal Procrustes
        R = torch.empty((self.rank, self.rank), device=W.device)
        torch.nn.init.orthogonal_(R)
        
        for _ in range(n_iter):
            # Step A: Project to Binary Vertices
            B = torch.sign(Z @ R)
            
            # Step B: Optimal Rotation (SVD of B.T @ Z)
            M = B.t() @ Z
            Phi, _, Psi_t = torch.linalg.svd(M, full_matrices=False)
            R = Psi_t.t() @ Phi.t() # R = V @ U.T

        # 4. Apply Aligned Rotation
        U_aligned = U @ R
        V_aligned = R.t() @ V
        
        # 5. Extract Scales (Robust due to bimodal distribution)
        h, l, g = self._extract_scales(U_aligned, V_aligned)
        return U_aligned, V_aligned, h, l, g
\end{lstlisting}


\end{document}